\documentclass[10pt,twocolumn,letterpaper]{article}

\usepackage{iccv}
\usepackage{times}
\usepackage{epsfig}

\usepackage[pagebackref=true,breaklinks=true,letterpaper=true,colorlinks,bookmarks=false]{hyperref}

\usepackage{amsmath,amssymb} % define this before the line numbering.
\usepackage{makecell}
\usepackage{graphics} % for pdf, bitmapped graphics files
\usepackage{colortbl}
\usepackage{empheq}
\usepackage{bm}
\usepackage{xcolor}
\usepackage{bbding} %首先在导言区调用bbding包
\usepackage{diagbox}
\usepackage[linesnumbered,ruled]{algorithm2e}
\usepackage{xcolor}
\newcommand{\cmark}{\ding{51}}%
\newcommand{\xmark}{\ding{55}}%
\usepackage[capitalize]{cleveref}
\usepackage{float}
\usepackage{booktabs}
\usepackage{multirow}
\usepackage{mathrsfs}
\usepackage[utf8]{inputenc}
\usepackage{pifont}
\usepackage{threeparttable}

\newcounter{RNum}
\renewcommand{\theRNum}{\arabic{RNum}}
\newcommand{\Remark}{\noindent\textbf{Remark}~\refstepcounter{RNum}\textbf{\theRNum}: }
\newcommand{\fref}[1]{Fig.~\ref{#1}}
\newcommand{\sref}[1]{Section~\ref{#1}}
\newcommand{\tref}[1]{Table~\ref{#1}}
\newcommand{\appref}[1]{Appendix~\ref{#1}}

\newcommand{\myparagraph}[1]{\noindent\textbf{#1}~}

\iccvfinalcopy % *** Uncomment this line for the final submission

\def\iccvPaperID{2191} % *** Enter the ICCV Paper ID here

\ificcvfinal\fi

\begin{document}

%%%%%%%%% TITLE
\title{PVT++: A Simple End-to-End Latency-Aware Visual Tracking Framework}

\author{Bowen Li$^{1, *}$, Ziyuan Huang$^{2, *}$, Junjie Ye$^{3}$, Yiming Li$^4$, Sebastian Scherer$^1$, Hang Zhao$^5$, Changhong Fu\textsuperscript{3, \ding{41}}
\and
% $^1$Robotics Institute, Carnegie Mellon University, USA\\
% $^2$Advanced Robotics Centre, National University of Singapore, Singapore\\
% $^3$School of Mechanical Engineering, Tongji University, China\\
% $^4$School of Mechanical and Aerospace Engineering, New York University, USA\\
% $^5$IIIS, Tsinghua University, China
$^1$Carnegie Mellon University, \ $^2$National University of Singapore, \ $^3$Tongji University\\
$^4$New York University, \ $^5$Tsinghua University\\
{\tt\small \{bowenli2, basti\}@andrew.cmu.edu, ziyuan.huang@u.nus.edu}\\
{\tt\small \{ye.jun.jie, changhongfu\}@tongji.edu.cn, yimingli@nyu.edu, hangzhao@tsinghua.edu.cn }
}

\maketitle
\ificcvfinal\thispagestyle{empty}\fi

%%%%%%%%% ABSTRACT
\begin{abstract}
   Visual object tracking is essential to intelligent robots. 
    Most existing approaches have ignored the online latency that can cause severe performance degradation during real-world processing.
    Especially for unmanned aerial vehicles (UAVs), where robust tracking is more challenging and onboard computation is limited, the latency issue can be fatal.
    In this work, we present a simple framework for end-to-end latency-aware tracking, \textit{i.e.}, end-to-end predictive visual tracking (PVT++). 
    Unlike existing solutions that naively append Kalman Filters after trackers, PVT++ can be jointly optimized, so that it takes not only motion information but can also leverage the rich visual knowledge in most pre-trained tracker models for robust prediction.
    Besides, to bridge the training-evaluation domain gap, we propose a relative motion factor, empowering PVT++ to generalize to the challenging and complex UAV tracking scenes.
    These careful designs have made the small-capacity lightweight PVT++ a widely effective solution.
    Additionally, this work presents an extended latency-aware evaluation benchmark for assessing an \textit{any-speed} tracker in the online setting. 
    Empirical results on a robotic platform from the aerial perspective show that PVT++ can achieve significant performance gain on various trackers and exhibit higher accuracy than prior solutions, largely mitigating the degradation brought by latency. 
    Our code is public at \url{https://github.com/Jaraxxus-Me/PVT_pp.git}.
\end{abstract}

%%%%%%%%% BODY TEXT
\section{Introduction}
\label{sec:intro}

    \begin{figure}[!t]
	\centering
	\includegraphics[width=1.03\columnwidth]{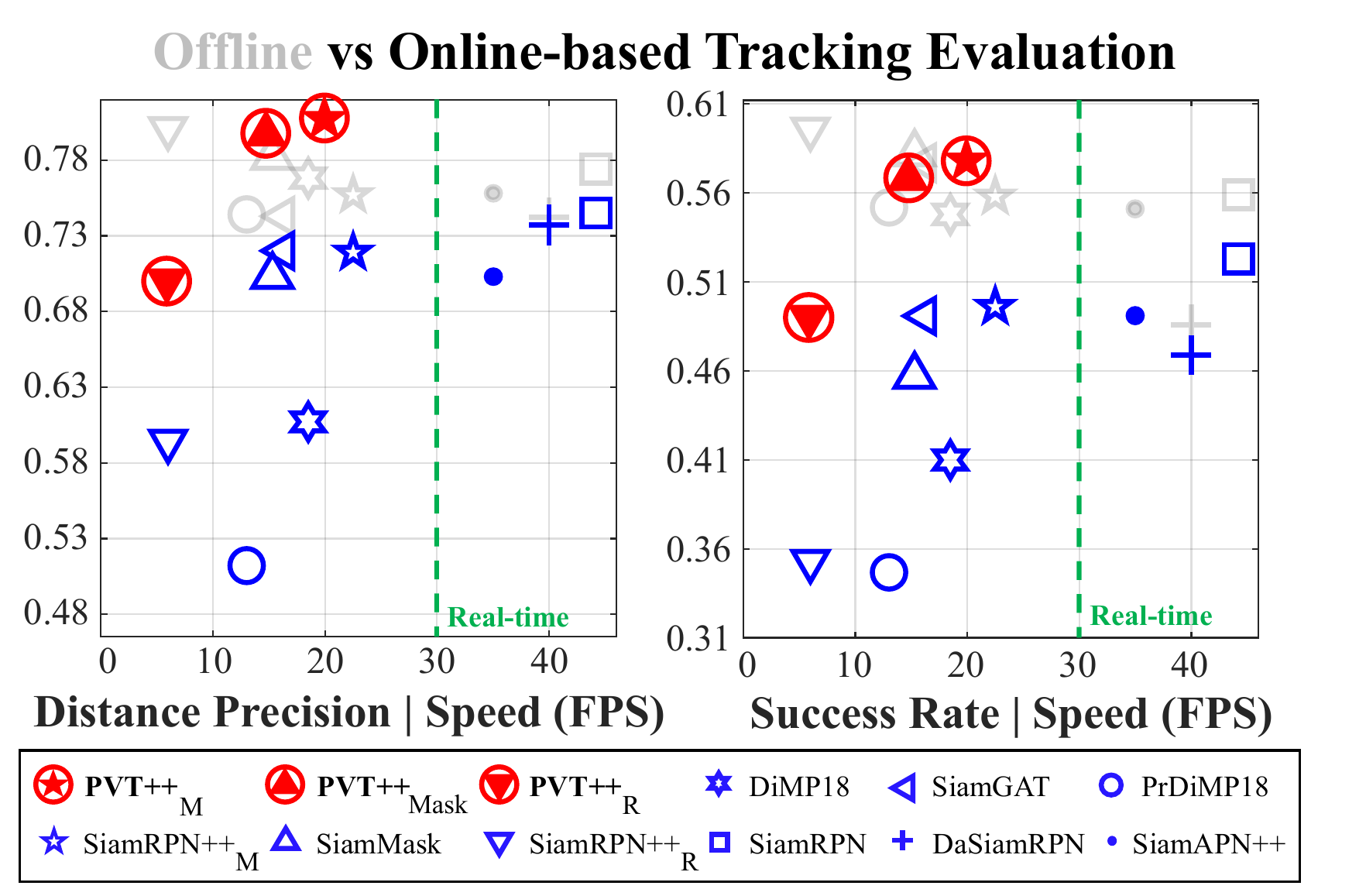}
        \vspace{-0.4cm}
        \caption{Distance precision and success rate of the trackers on UAVDT dataset \cite{Du2018UAVDT}. Compared with {\color{gray}{\textbf{offline}}} evaluation, the trackers suffer a lot from their onboard latency in the \textbf{online} setting (30 frames/s (FPS)). Coupled with \textbf{PVT++}, the predictive trackers achieve significant performance gain with very little extra latency, obtaining on par or better results than the {\color{gray}{\textbf{offline}}} setting.
	}
        \vspace{-0.4cm}
 
	\label{fig:star}
    \end{figure}

    \begin{figure*}[!t]
	\centering
	\includegraphics[width=1.85\columnwidth]{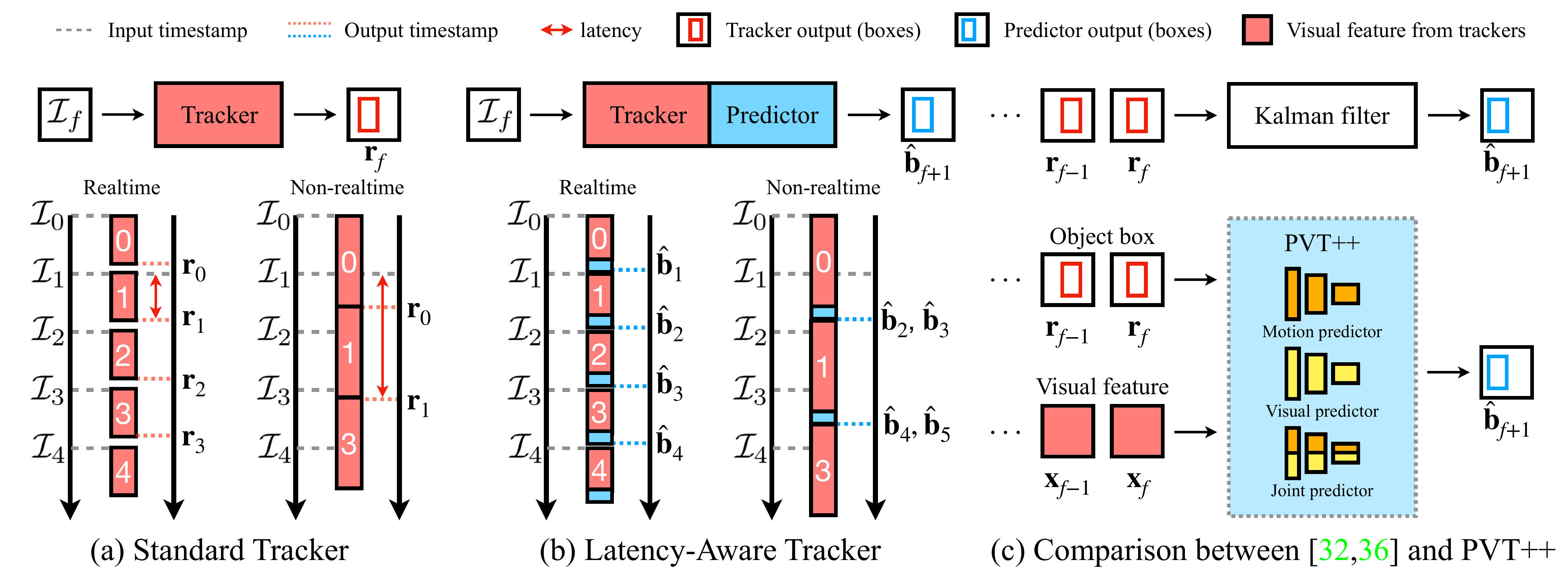}
	\caption{(a) Standard tracker suffers from onboard latency (height of the red boxes). Hence, its result lags behind the world, \textit{i.e.}, $\mathbf{r}_f$ is always obtained after $\mathcal{I}_f$ on the timestamp.
	(b) Latency-aware trackers introduce predictors to compensate for the latency, which predict the word state, $\hat{\mathbf{b}}_{f+1}$, when finishing the processed frame. 
	(c) Compared with prior KF-based solutions~\cite{Li2020PredictiveVT, Li2020TowardsSP}, our end-to-end framework for latency-aware tracking PVT++ leverages both motion and visual feature for prediction.
	}
        \vspace{-0.4cm}
	\label{fig:preliminary}
    \end{figure*}

    Visual object tracking\footnote{We focus on single object tracking in this work.} is fundamental for many robotic applications like navigation \cite{Nishida2018Navigation}, cinematography \cite{Bonatti2019Cinema}, and multi-agent cooperation \cite{Chen2020Cooperation}. 
    Most existing trackers are developed and evaluated under an \textit{offline} setting~\cite{Li2020AutoTrackTH,Huang2019LearningAR,Li2018HighPV,Li2019SiamRPNEO,Cao2021HiFT,Cao2022TCTrack}, where the trackers are assumed to have zero processing time. 
    However, in real-world deployment, the online latency caused by the trackers' processing time cannot be ignored, since the world would have already changed when the trackers finish processing the captured frame. 
    In particular, with limited onboard computation, this issue is more critical in the challenging unmanned aerial vehicle (UAV) tracking scenes \cite{Fu2022GRSM,Li2020AutoTrackTH,Cao2021APN}.
    As shown in \fref{fig:star}, compared with \textit{offline} setting ({\color{gray}{gray}} markers), the latency can cause severe performance degradation during \textit{online} processing ({colored markers}).
    If not handled well, this can easily lead to the failure of robotic applications such as UAV obstacle avoidance~\cite{Aguilar2019Avoid} and self-localization~\cite{Ye2021TIE}.

    To be more specific, the latency hurts online tracking due to: (1) The tracker outputs are always outdated, so there will be mismatch between the tracker result and world state. (2) The trackers can only process the latest frame, so that the non-real-time ones may skip some frames, which makes object motion much larger (see Fig.~\ref{fig:preliminary}(a) right).
    
    The existence of the latency in real-world applications calls for trackers with prediction capabilities, \textit{i.e.,} predictive trackers. 
    While a standard tracker yields the objects' location in the input frame (\textit{i.e.}, when it \textit{starts} processing the input frame, as in Fig.~\ref{fig:preliminary}(a)), a predictive tracker predicts where the objects could be when it \textit{finishes} processing the input frame, as illustrated in Fig~\ref{fig:preliminary}(b). 
    
    Existing solutions~\cite{Li2020PredictiveVT,Li2020TowardsSP} directly append a Kalman filter (KF)~\cite{kalman1960} after trackers to estimate the potential object's location based on its motion model (see Fig~\ref{fig:preliminary}(c)).
    However, the rich and readily available visual knowledge from trackers is primarily overlooked, including the object's appearance and the surrounding environments, which can be naturally exploited to predict the objects' future paths~\cite{rudenko2020human}.
    
    To this end, we present a simple framework PVT++ for end-to-end predictive visual tracking. 
    Composed of a tracker and a predictor, PVT++ is able to convert most off-the-shelf trackers into effective predictive trackers.
    Specifically, to avoid extra latency brought by the predictor, we first design a lightweight network architecture, consisting of a feature encoder, temporal interaction module, and predictive decoder, that leverage both historical motion information and visual cues.
    By virtue of joint optimization, such a small-capacity network can directly learn from the visual representation provided by most pre-trained trackers for an efficient and accurate motion prediction, as in Fig.~\ref{fig:preliminary}(c). 
    However, learning this framework is non-trivial due to the training-evaluation domain gap in terms of motion scales.
    To solve this, we develop a relative motion factor as training objective, so that our framework is independent of the motion scales in training data and can generalize well to the challenging aerial tracking scenes. 
    The integration of lightweight structure and training strategy yields an effective, efficient, and versatile solution.

    Beyond methodology, we found that the existing latency-aware evaluation benchmark (LAE)~\cite{Li2020PredictiveVT} is unable to provide an effective latency-aware comparison for real-time trackers, since it evaluates the result for each frame as soon as it is given. In this case, the latency for any real-time trackers is one frame.
    Hence, we present an extended latency-aware evaluation benchmark (e-LAE) for \textit{any-speed} trackers. 
    Evaluated with various latency thresholds, real-time trackers with different speeds can be distinguished.
    
    Empirically, we provide a more general, comprehensive, and practical aerial tracking evaluation for state-of-the-art trackers using our new e-LAE.
    Converting them into predictive trackers, PVT++ achieves up to \textbf{60}\% improvement under the \textit{online} setting. As shown in \fref{fig:star}, powered by PVT++, the predictive trackers can achieve comparable or better results than the \textit{offline} setting.
    Extensive experiments on multiple tracking models \cite{Li2019SiamRPNEO,Wang2019Mask,Guo2021SiamGAT} and datasets \cite{Mueller2016UAV123,Du2018UAVDT,Li2017DTB70} show that PVT++ works generally for latency-aware tracking, which, to the best of our knowledge, is also the first end-to-end framework for \textit{online} visual tracking.

    % We summarize our contribution as:
    % \begin{itemize}
    %     \item We propose end-to-end latency-aware visual tracking framework, which learns the rich visual knowledge from trackers with extraordinary onboard efficiency.
    %     \item We discover the relative motion factor as training objective, which helps bridge the training-evaluation domain gap and contributes to a generalizable method.
    %     \item We develop a more general latency-aware benchmark and conduct exhaustive evaluation of various trackers.
    %     \item Extensive experiments on robotics platform and real-world tests validate our widely-effective framework.
    % \end{itemize}

%-------------------------------------------------------------------------
\section{Related Work}
\label{sec:related}

\label{sec:eLAE}
% Fig. 2, (a) previous benchmark
% (b) our new benchmark
\begin{figure*}[!t]
	\centering
	\includegraphics[width=1.7\columnwidth]{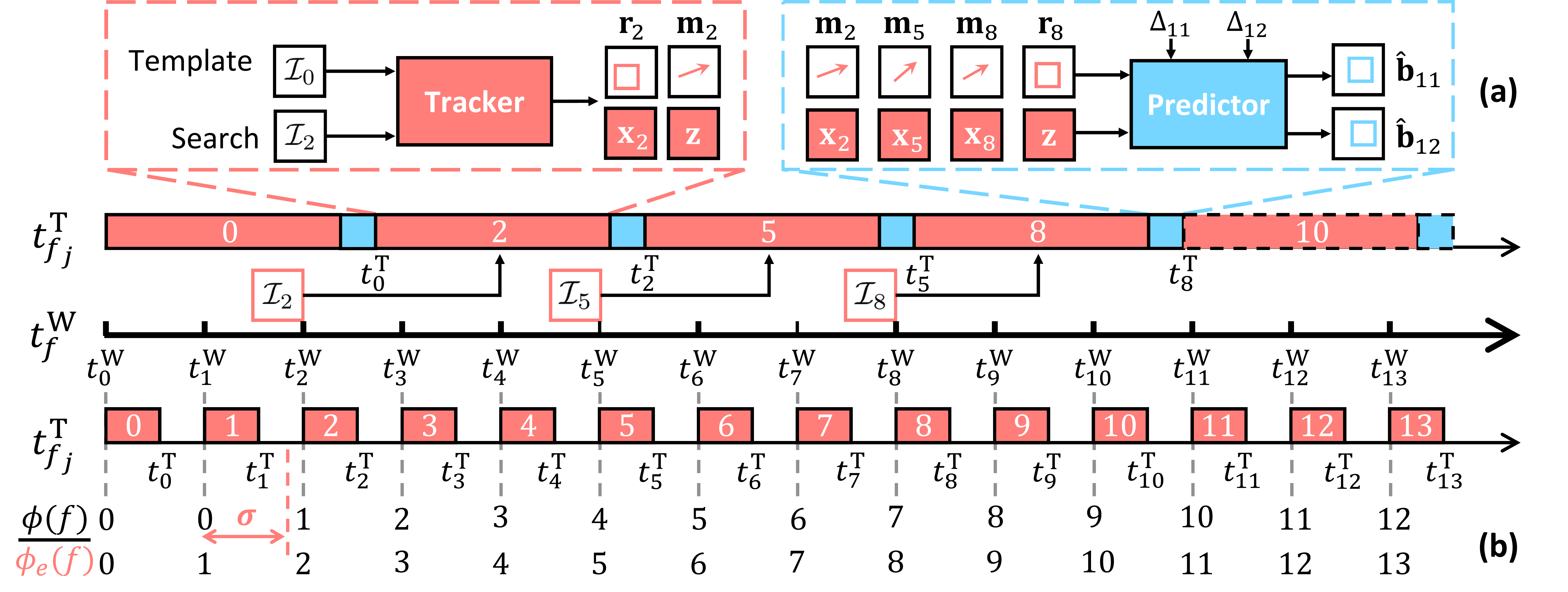}
  \vspace{-0.2cm}
	\caption{(a) Framework overview of PVT++ for a non-real-time tracker. The tracker has processed frame $0, 2, 5, 8$ and obtained corresponding motions $\mathbf{m}$ and visual features $\mathbf{x}, \mathbf{z}$. The predictor needs to predict future box $\hat{\mathbf{b}}_{11}$, $\hat{\mathbf{b}}_{12}$ based on tracker result $\mathbf{r}_8$. (b) Comparison between LAE ($\phi(f)$)~\cite{Li2020PredictiveVT} and our e-LAE ($\phi_e(f)$). For real-time trackers, the mismatch between output and input frames will always be one in LAE ($\phi(f)-f\equiv1$) regardless of the trackers' various latency. Differently, e-LAE introduces permitted latency thresholds $\sigma\in[0,1)$, which effectively distinguishes the latency difference of distinct models.}
 \vspace{-0.4cm}
	\label{fig:method}
\end{figure*}

\subsection{Visual Tracking and its Aerial Applications}

    Visual trackers basically fall into two paradigms, respectively based on discriminative correlation filters~\cite{Bolme2010MOSSE,Henriques2015KCF,Danelljan2015ICCV,Danelljan2017CVPR} and Siamese networks~\cite{Bertinetto2016SiamFC,Li2018HighPV,Zhu2018DistractorawareSN,Li2019SiamRPNEO,Guo2020SiamCAR,Xu2020SiamFC++}. 
    Compared with general scenarios, aerial tracking is more challenging due to large motions and limited onboard computation resources. 
    Hence, for efficiency, early approaches focus on correlation filters~\cite{Li2020AutoTrackTH,Huang2019LearningAR,Li2020KeyfilterAwareRU,Li2020TrainingSetDF}. 
    Later, the development of onboard computation platforms facilitates more robust and applicable Siamese network-based approaches~\cite{Cao2021APN,Cao2021APN++,Cao2021HiFT,Cao2022TCTrack}.
    
    Most of them are designed under \textit{offline} settings, ignoring the online latency onboard UAVs, which can lead to severe accuracy degradation. We aim to solve the more practical and challenging latency-aware perception problem, which goes beyond tracking and is often neglected.

\subsection{Latency-Aware Perception}

    Latency of perception systems is first studied in~\cite{Li2020TowardsSP}, which introduces a baseline based on the Kalman-filter~\cite{kalman1960} to compensate for the online latency of object detectors. Inspired by this, \cite{Yang2022StreamYolo} converts a real-time detector into a latency-aware one. 
    More recent work \cite{sela2022context} also leverage context scenario properties to select best configurations.
    While most previous work aims at object detection, 
    \cite{Li2020PredictiveVT} is more related to us in the field of visual tracking, which introduces dual KFs \cite{kalman1960} similar to~\cite{Li2020TowardsSP}.
    Overall, most existing works didn't address latency-aware tracking through a deeply coupled framework. In this work, we target aerial tracking and present an end-to-end structure.

\subsection{Tracking with/by Prediction}
% TODO
    Integrating prediction into trackers has been widely adopted for robust perception \cite{liang2020PnPNet,rudenko2020human,luo2018fast,fernando2018trackingbypred,liang2018real,liu2020object}. 
    
    Some previous works~\cite{liang2018real,liu2020object} introduce a predictor to correct the tracker results under object occlusion, which potentially share similar design with PVT++. 
    However, they are still developed for \textit{offline} tracking, \textit{i.e.}, the predictor works only for current frame. 
    Yet predicting the future state for aerial tracking is non-trival due to the training-evaluation gap. PVT++ aims to address this gap via a relative motion factor, thus working beyond \cite{liang2018real,liu2020object}.
    
    Others~\cite{liang2020PnPNet,rudenko2020human,luo2018fast,fernando2018trackingbypred} focus on trajectory prediction after tracking. While PVT++ predicts both the trajectory and the object scale. 
    Moreover, as a result of fast camera motion and viewpoint change, UAV tracking scenes are much more complex than the ordered environments in autonomous driving \cite{liang2020PnPNet,luo2018fast}. 
    PVT++ discovers the valuable visual knowledge in pre-trained trackers, yielding its capability for such challenging condition with small capacity.

\subsection{Visual Tracking Benchmarks} 

    Various benchmarks are built for large-scale tracking evaluation
    ~\cite{Fan2019LaSOTAH,Mller2018TrackingNetAL,Huang2019GOT10kAL,Dunnhofer2020IsFP,Liu2020PTBTIRAT,Mueller2016UAV123,Li2021ADTrack,Ye2022UDAT} with different challenges such as first-person perspective \cite{Dunnhofer2020IsFP}, aerial scenes \cite{Mueller2016UAV123}, illumination conditions \cite{Li2021ADTrack,Ye2022UDAT}, and thermal infrared inputs \cite{Liu2020PTBTIRAT}. Since they all adopt \textit{offline} evaluation, the influence of the trackers' latency is ignored. A recent benchmark targets online evaluation~\cite{Li2020PredictiveVT}, but it falls short in real-time trackers and we aim to improve it in this work.

\section{Preliminary}

% Fig. 4, architecture for 3 models
\begin{figure*}[!t]
	\centering
	\includegraphics[width=1.8\columnwidth]{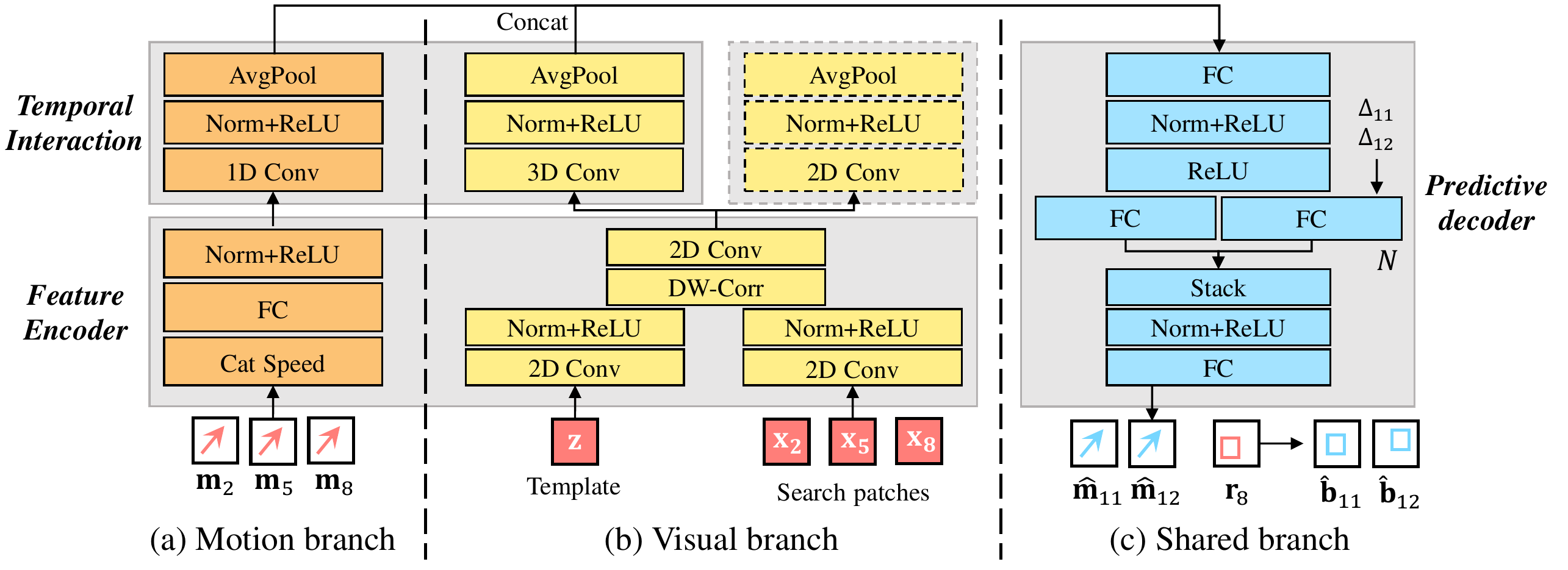}
        \vspace{-0.2cm}
	\caption{Detailed model structure of the predictor modules in PVT++. The models shares similar architecture, \textit{i.e.}, feature encoder, temporal interaction, and predictive decoder. We present the motion branch, visual branch, and share decoding branch in (a), (b), and (c), respectively. Note that the dashed blocks denote auxiliary branch, which only exists in training. The input and output are in correspondence to the case in \fref{fig:method} (a).}
        \vspace{-0.4cm}
	\label{fig:structure}
\end{figure*}

    We first introduce the latency-aware tracking task here.
    The input is an image sequence broadcasting with a certain framerate $\kappa$, denoted as $(\mathcal{I}_f, t^{\mathrm{W}}_f), f\in \{0,1,2,\cdots\}$, where $t^{\mathrm{W}}_f=\frac{f}{\kappa}$ is the world timestamp and $f$ is the frame index. 
    Provided with the ground truth box $\mathbf{b}_0= [x_0, y_0, w_0, h_0]$ at initial $0$-th frame, the tracker estimates the boxes in the following frames. Detailed notation table see \appref{app:notation}.

	\myparagraph{Inference. }During inference, the tracker finds the \textit{latest} frame to process when finishing the previous one. 
	Due to the latency, for the $j$-th frame that the tracker processes, its index $j$ may differ from its frame index $f_j$ in the image sequence. 
	The frame to be processed (frame $f_j$) is determined by the tracker timestamp $t^{\mathrm{T}}_{f_{j-1}}$ when the model finishes frame $f_{j-1}$ as follows:
	\begin{equation}
	\small
    f_j = 
	    \left\{
    	    \begin{array}{crl}
    	    0 & , & j=0 \\
    	    \mathop{\arg\max}_{f} t^{\mathrm{W}}_f\le t^{\mathrm{T}}_{f_{j-1}} & , & \mathrm{others}
            \end{array}
        \right.~.
	\end{equation}
 
	With the frame index $f_j$, the tracker processes frame $\mathcal{I}_{f_j}$ to obtain the corresponding box $\mathbf{r}_{f_j} = [x_{f_j}, y_{f_j}, w_{f_j}, h_{f_j}]$, forming the raw result of the tracker on the frame $(\mathbf{r}_{f_j}, t^{\mathrm{T}}_{f_j})$.
	Since tracker may be non-real-time, input frame ids ${f_j}, j\in \{0,1,2,\cdots\}$ may not be consecutive numbers. 
	For example, in \fref{fig:method} (a), considering a non-real-time tracker, the processed frames are $f_j=0, 2, 5, 8, \cdots$.
	
	\myparagraph{Evaluation. }Latency-aware evaluation (LAE)~\cite{Li2020PredictiveVT} compares the ground-truth $\mathbf{b}_f$ in frame $\mathcal{I}_f$ with the \textit{latest} result $\hat{\mathbf{b}}_f$ from the tracker at $t^{\mathrm{W}}_f$ for evaluation.
	For standard trackers, the latest result $\hat{\mathbf{b}}_f$ to be compared with the ground-truth is obtained as $\hat{\mathbf{b}}_f = \mathbf{r}_{\phi{(f)}}$, where $\phi{(f)}$ is defined as:
	\begin{equation}\label{eqn:eva}
	\small
	    \phi{(f)} = 
	    \left\{
	    \begin{array}{crl}
	    0 & , & t^{\mathrm{W}}_f < t^{\mathrm{T}}_{f_0} \\
	    \mathop{\arg\max}_{f_j} t^{\mathrm{T}}_{f_j}\le t^{\mathrm{W}}_f & , & \mathrm{others}
        \end{array}\right.~.
	\end{equation}
 
	For instance, in \fref{fig:method} (b), LAE compares the ground truth $\mathbf{b}_3$ with the raw tracker result $\mathbf{r}_2$.
	
\section{Extended Latency-Aware Benchmark}

    Existing latency-aware evaluation \cite{Li2020TowardsSP,Li2020PredictiveVT} adopt Eq.~(\ref{eqn:eva}) to match the raw output $(\mathbf{r}_{f_j}, t^{\mathrm{T}}_{f_j})$ to every input frame $f$.
    However, such a policy fails to reflect the latency difference among real-time trackers.
    As shown in \fref{fig:method}, since the real-time methods is faster than frame rate, every frame will be processed, \textit{i.e.}, $[f_0, f_1, f_2, \cdots]=[0, 1, 2, \cdots]$. In this case, the \textit{latest} results will always be from the previous one frame, \textit{i.e.}, using Eq.~(\ref{eqn:eva}), $\phi(f)\equiv f-1$.
    Differently, we extend Eq.~(\ref{eqn:eva}) to:
	\begin{equation}\label{eqn:elae}
	    \small
	    \phi{(f)}_{\mathrm{e}} = 
	    \left\{
	    \begin{array}{crl}
	    0 & , & t^{\mathrm{W}}_f < t^{\mathrm{T}}_{f_0} \\
	    \mathop{\arg\max}_{f_j} t^{\mathrm{T}}_{f_j}\le t^{\mathrm{W}}_f + \sigma & , & \mathrm{others}
        \end{array}\right.~,
	\end{equation}
    where $\sigma\in[0, 1)$ is the variable permitted latency. 
    Under e-LAE, $\phi{(f)}_{\mathrm{e}}$ can be $f-1$ or $f$ for real-time trackers depending on $\sigma$. For instance, $\phi{(f)}_{\mathrm{e}}$ would turn from $f-1$ to $f$ at larger $\sigma$ for slower real-time trackers.
    This extension distinguishes different real-time trackers (see \sref{sec:elae}).

\section{Predictive Visual Tracking}

\label{sec:method}
    Because of the unavoidable latency introduced by the processing time, there is always a mismatch between $\phi(f)$ (or $\phi{(f)}_{\mathrm{e}}$) and $f$ (when $\sigma$ is small), where $\phi(f)$ is always smaller than $f$, \textit{i.e.,} $\phi(f)<f, f>0$. To compensate for the mismatch, we resort to predictive trackers that predicts possible location of the object in frame $f$. For the evaluation of $f$-th frame, prior attempts \cite{Li2020TowardsSP, Li2020PredictiveVT} adopt traditional KF \cite{kalman1960} to predict the result based on the raw tracking result $\mathbf{r}_{\phi{(f)}}$ in $\mathcal{I}_\phi{(f)}$~\cite{Li2020TowardsSP}, \textit{i.e.,} $\hat{\mathbf{b}}_f=\text{KF}(\mathbf{r}_{\phi{(f)}})$.
    Since previous work \cite{Li2020TowardsSP, Li2020PredictiveVT} are not learnable, neither existing large-scale datasets nor the visual feature are leveraged. 
    Differently, our predictive visual tracking framework PVT++ aims for an end-to-end predictive tracker, which takes both the historical motion and visual features for a more robust and accurate prediction. 
    Note that we use $\hat{\cdot}$ to represent the prediction (results for evaluation) and others are from the tracker output or ground-truth in the following subsections.
	
\subsection{General Framework}

% Fig. 6, overall evaluation results
\begin{figure*}[!t]
	\centering
	\includegraphics[width=2.1\columnwidth]{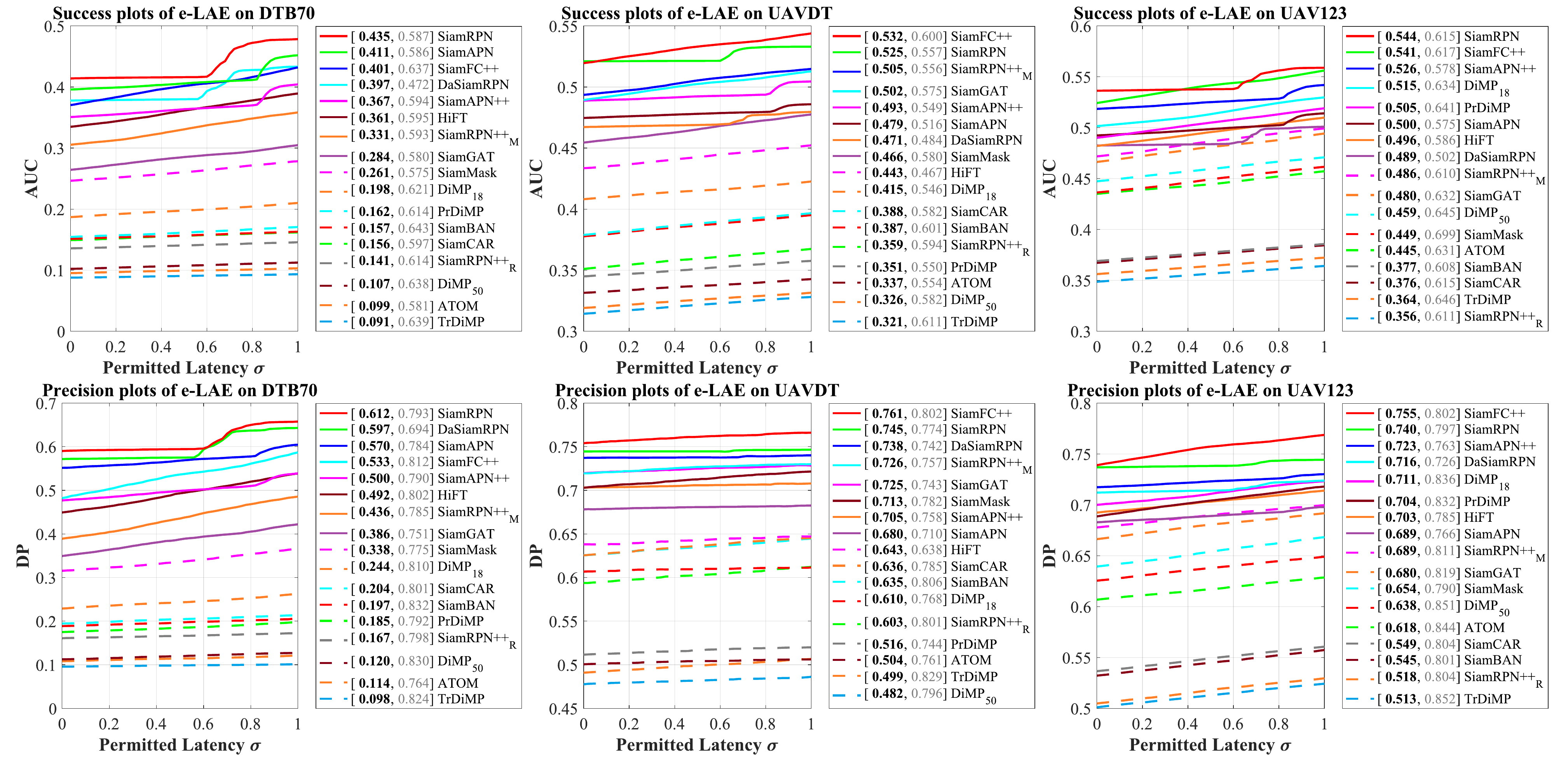}
 \vspace{-0.4cm}
	\caption{The performance of the SOTA trackers in authoritative UAV tracking benchmarks under our e-LAE benchmark. 
	We report [online mAUC and mDP, {\color{gray}offline AUC and DP}] in the legend.
	All trackers struggle to overcome onboard latency in online tracking.}
        \vspace{-0.4cm}
	\label{fig:elae}
\end{figure*}

    As in \fref{fig:method} (a), PVT++ consists of a tracker $\mathcal{T}$ and a predictor $\mathcal{P}$. 
    For the $f$-th frame at world time $t^\mathrm{W}_f$, the latest result from the tracker is ${\mathbf{r}}_{\phi{(f)}}$ obtained from frame $\mathcal{I}_{\phi{(f)}}$, \textit{i.e.}, $\mathbf{r}_{\phi{(f)}}=\mathcal{T}(\mathbf{x}_{\phi{(f)}},\mathbf{z})$, where $\mathbf{x}_{\phi{(f)}}$ is the search feature from $\mathcal{I}_{\phi{(f)}}$ and $\mathbf{z}$ is the template feature.

    After this, the predictor $\mathcal{P}$ takes input from the information generated during tracking of the $k$ past frames (including $\mathcal{I}_{\phi(f)}$), denoted as $\text{Input}_{\phi{(f)}}$, and predict the position offset normalized by object's scale, \textit{i.e.,} motion $\hat{\mathbf{m}}_{f} = [\frac{\Delta_{\hat{x}}(f)}{w_{\phi{(f)}}}, \frac{\Delta_{\hat{y}}(f)}{h_{\phi{(f)}}}, \mathrm{Log}(\frac{\hat{w}_{f}}{w_{\phi(f)}}), \mathrm{Log}(\frac{\hat{h}_{f}}{h_{\phi{(f)}}})]$, where $\Delta_{\hat{x}}(f)$ and $\Delta_{\hat{y}}(f)$ denote the predicted box center distance between the $f$-th and $\phi(f)$-th frame. 
	$w_{\phi(f)}$ and $h_{\phi(f)}$ are the tracker's output box scale in frame $\phi(f)$ and $\hat{w}_{f}$, $\hat{h}_{f}$ are the predicted scale in $f$-th frame.
	With the raw output $\mathbf{r}_{\phi{(f)}}$ at $\phi(f)$ and the motion $\hat{\mathbf{m}}_{f}$ from $\mathcal{I}_{\phi(f)}$ to the $f$-th frame, the predicted box $\hat{\mathbf{b}}_f$ can be easily calculated.
    
    \myparagraph{Relative Motion Factor:}Due to the large domain gap between the training \cite{Russakovsky2015VID} and evaluation \cite{Li2017DTB70} in terms of the absolute motion scale, we find directly using the absolute motion value ${\hat{\mathbf{m}}}_{f}$ as the objective can result in poor performance (see \tref{tab:abla}). 
    Therefore, we define the output of predictor $\mathcal{P}$ to be the relative motion factor based on the average moving speed $\mathbf{p}_{f_j}$ from the past $k$ frames, which we find is easier to generalize after training:\begin{equation}\label{eqn:general}
        \small
        \hat{\mathbf{m}}_{f} = \mathcal{P}\big(\text{Input}_{\phi{(f)}}, \Delta_f\big)\odot{\mathbf{p}_{f_j}}\ ,\quad
	    \mathbf{p}_{f_j} = \frac{1}{k}\sum_{i=j-k+1}^{j}\frac{\mathbf{m}_{f_i}}{\Delta_{f_i}},
    \end{equation}
    where $\Delta_{f} = f - \phi(f)$ denotes the frame interval between current and target frame, and $f_j$ is the latest processed frame, \textit{e.g.}, $\phi(f)$. $\Delta_{f_i} = f_{i} - f_{i-1}$ denotes the frame interval between $(i-1)$ and $i$-th processed frame. $\odot$ indicates element-wise multiplication. $\mathbf{m}_{f_{i}}$ is the normalized input motion defined as $\mathbf{m}_{f_{i}} = [\frac{\Delta_x({f_{i}})}{w_{f_{i-1}}}, \frac{\Delta_y({f_{i}})}{h_{f_{i-1}}}, \mathrm{Log}(\frac{w_{f_{i}}}{w_{f_{i-1}}}), \mathrm{Log}(\frac{h_{f_{i}}}{h_{f_{i-1}}})]$, where $\Delta_x({f_{i}}) = x_{f_{i}} - x_{f_{i-1}}$ and $\Delta_y({f_{i}}) = y_{f_{i}} - y_{f_{i-1}}$ are the distance from tracker results $\mathbf{r}_{f_i}$ and $\mathbf{r}_{f_{i-1}}$. Such design has made PVT++ agnostic to the specific motion of a dataset, which is crucial for its generalization capability.
    
    We next present the predictor of PVT++ step by step as motion-based $\mathcal{P}_{\mathrm{M}}$, visual-appearance-based $\mathcal{P}_{\mathrm{V}}$ and multi-modal-based $\mathcal{P}_{\mathrm{MV}}$.
    All the predictors share the same training objective (Eq.~\eqref{eqn:general}) and a similar structure, consisting of feature encoding, temporal interaction, and predictive decoding as in \fref{fig:structure}.
    In practice, a predictor may need to predict $N$ results, depending on the tracker's latency. 
    % PVT++ realize this by $N$ parallel fully connected (FC) layers in \fref{fig:structure} (c).
    
    % \Remark Unlike KFs \cite{Li2020TowardsSP,Li2020PredictiveVT} need to do $N$ times prediction, PVT++ output $N$ predictions in one forward pass, which is more efficient.  
    
\subsection{Motion-based Predictor}

    The motion-based predictor $\mathcal{P}_{\mathrm{M}}$ only relies on the past motion, \textit{i.e.,} $\text{Input}_{\phi(f)}=\mathbf{m}_{f_{j-k+1}:f_j}$,
    \begin{equation}
        \small
        \hat{\mathbf{m}}_{f,\mathrm{M}} = \mathcal{P}_{\mathrm{M}}\big(\mathbf{m}_{f_{j-k+1}:f_j}, \Delta_f\big)\odot{\mathbf{p}_{f_j}}\ ,
    \end{equation}
    where $\mathbf{m}_{f_{j-k+1}:f_j}=[\mathbf{m}_{f_{j-k+1}}, \cdots, \mathbf{m}_{f_{j}}]\in\mathbb{R}^{k\times 4}$.

	The detailed model structure of the motion predictor $\mathcal{P}_\mathrm{M}$ is presented in \fref{fig:structure}(a). 
	For pre-processing, the motion data $\mathbf{m}_{f_{j-k+1}}, \cdots, \mathbf{m}_{f_{j}}$ are first concatenated.
	Then we apply a fully connected (FC) layer with non-linearity for feature encoding and a 1D convolution followed by activation and global average pooling to obtain the temporally interacted motion feature. 
	In the predictive decoding head, a share FC layer with non-linearity is used for feature mapping. $N$ independent FCs map the feature to $N$ future latent spaces. Finally, the latency features are stacked and transformed to $4$ dimension output using a shared FC.

	For training, we adopt $\mathcal{L}_1$ loss between prediction and ground-truth $\mathcal{L}_{\mathrm{M}} = \mathcal{L}_1(\hat{\mathbf{m}}_{f,M}, \mathbf{m}_{f})$.
	    
\subsection{Visual Appearance-based Predictor}

% PVT +
\begin{table*}[!t]
	\centering
	\setlength{\tabcolsep}{0.4mm}
	\fontsize{6.5}{7.5}\selectfont
	\caption{The effect of PVT++ on the four SOTA trackers with different inference speeds and backbones. Our models work generally for different tracker structures and  can achieve up to \textbf{60}\% performance gain. The best scores are marked out in \colorbox[rgb]{ .806,  .802,  .802}{gray} for clear reference. We present some qualitative visualization in \appref{app:visual} and the supplementary video.}
	\begin{threeparttable}
    \begin{tabular}{cc|cccc|cccc|cccc|cccc}
    \toprule[1.5pt]
          & Dataset & \multicolumn{4}{c|}{DTB70}    & \multicolumn{4}{c|}{UAVDT}    & \multicolumn{4}{c|}{UAV20L}   & \multicolumn{4}{c}{UAV123} \\
    Tracker & PVT++  & \multicolumn{2}{c}{AUC@La0$_{\Delta\%}$} & \multicolumn{2}{c|}{DP@La0$_{\Delta\%}$} & \multicolumn{2}{c}{AUC@La0$_{\Delta\%}$} & \multicolumn{2}{c|}{DP@La0$_{\Delta\%}$} & \multicolumn{2}{c}{AUC@La0$_{\Delta\%}$} & \multicolumn{2}{c|}{DP@La0$_{\Delta\%}$} & \multicolumn{2}{c}{AUC@La0$_{\Delta\%}$} & \multicolumn{2}{c}{DP@La0$_{\Delta\%}$} \\
    \midrule
    \multicolumn{1}{c}{\multirow{4}[2]{*}{\makecell[c]{SiamRPN++$_{\mathrm{M}}$\\($\sim$22FPS)\\(MobileNet \cite{sandler2018mobilenetv2})}}} & N/A   & \multicolumn{2}{c}{0.305$_{+0.00}$} & \multicolumn{2}{c|}{0.387$_{+0.00}$} & \multicolumn{2}{c}{0.494$_{+0.00}$} & \multicolumn{2}{c|}{0.719$_{+0.00}$} & \multicolumn{2}{c}{0.448$_{+0.00}$} & \multicolumn{2}{c|}{0.619$_{+0.00}$} & \multicolumn{2}{c}{0.472$_{+0.00}$} & \multicolumn{2}{c}{0.678$_{+0.00}$} \\
          & $\mathcal{P}_\mathrm{M}$ & \multicolumn{2}{c}{0.385$_{+26.2}$} & \multicolumn{2}{c|}{0.523$_{+35.1}$} & \multicolumn{2}{c}{0.529$_{+7.10}$} & \multicolumn{2}{c|}{0.745$_{+3.60}$} & \multicolumn{2}{c}{0.481$_{+7.40}$} & \multicolumn{2}{c|}{0.647$_{+4.50}$} & \multicolumn{2}{c}{0.537$_{+13.8}$} & \multicolumn{2}{c}{0.737$_{+8.70}$} \\
          & $\mathcal{P}_\mathrm{V}$ & \multicolumn{2}{c}{0.352$_{+15.4}$} & \multicolumn{2}{c|}{0.472$_{+22.0}$} & \multicolumn{2}{c}{0.564$_{+14.2}$} & \multicolumn{2}{c|}{0.799$_{+11.1}$} & \multicolumn{2}{c}{0.488$_{+8.90}$} & \multicolumn{2}{c|}{0.675$_{+9.00}$} & \multicolumn{2}{c}{0.504$_{+6.80}$} & \multicolumn{2}{c}{0.703$_{+3.70}$} \\
          & $\mathcal{P}_\mathrm{MV}$ & \multicolumn{2}{c}{\cellcolor[rgb]{ .806,  .802,  .802}0.399$_{+30.8}$} & \multicolumn{2}{c|}{\cellcolor[rgb]{ .806,  .802,  .802}0.536$_{+38.5}$} & \multicolumn{2}{c}{\cellcolor[rgb]{ .806,  .802,  .802}0.576$_{+16.6}$} & \multicolumn{2}{c|}{\cellcolor[rgb]{ .806,  .802,  .802}0.807$_{+12.2}$} & \multicolumn{2}{c}{\cellcolor[rgb]{ .806,  .802,  .802}0.508$_{+13.4}$} & \multicolumn{2}{c|}{\cellcolor[rgb]{ .806,  .802,  .802}0.697$_{+12.6}$} & \multicolumn{2}{c}{\cellcolor[rgb]{ .806,  .802,  .802}0.537$_{+13.8}$} & \multicolumn{2}{c}{\cellcolor[rgb]{ .806,  .802,  .802}0.741$_{+9.30}$} \\
    \midrule
    \multicolumn{1}{c}{\multirow{4}[2]{*}{\makecell[c]{SiamRPN++$_{\mathrm{R}}$\\($\sim$6FPS)\\(ResNet50 \cite{He2015ResNet})}}} & N/A   & \multicolumn{2}{c}{0.136$_{+0.00}$} & \multicolumn{2}{c|}{0.159$_{+0.00}$} & \multicolumn{2}{c}{0.351$_{+0.00}$} & \multicolumn{2}{c|}{0.594$_{+0.00}$} & \multicolumn{2}{c}{0.310$_{+0.00}$} & \multicolumn{2}{c|}{0.434$_{+0.00}$} & \multicolumn{2}{c}{0.349$_{+0.00}$} & \multicolumn{2}{c}{0.505$_{+0.00}$} \\
          & $\mathcal{P}_\mathrm{M}$ & \multicolumn{2}{c}{0.199$_{+46.3}$} & \multicolumn{2}{c|}{\cellcolor[rgb]{ .806,  .802,  .802}0.258$_{+62.3}$} & \multicolumn{2}{c}{0.449$_{+27.9}$} & \multicolumn{2}{c|}{0.684$_{+15.2}$} & \multicolumn{2}{c}{0.404$_{+30.3}$} & \multicolumn{2}{c|}{0.560$_{+29.0}$} & \multicolumn{2}{c}{0.442$_{+26.6}$} & \multicolumn{2}{c}{\cellcolor[rgb]{ .806,  .802,  .802}0.627$_{+24.2}$} \\
          & $\mathcal{P}_\mathrm{V}$ & \multicolumn{2}{c}{0.179$_{+31.6}$} & \multicolumn{2}{c|}{0.225$_{+41.5}$} & \multicolumn{2}{c}{0.403$_{+14.8}$} & \multicolumn{2}{c|}{0.665$_{+12.0}$} & \multicolumn{2}{c}{0.398$_{+28.4}$} & \multicolumn{2}{c|}{0.548$_{+26.3}$} & \multicolumn{2}{c}{0.398$_{+14.0}$} & \multicolumn{2}{c}{0.559$_{+10.7}$} \\
          & $\mathcal{P}_\mathrm{MV}$ & \multicolumn{2}{c}{\cellcolor[rgb]{ .806,  .802,  .802}0.205$_{+50.7}$} & \multicolumn{2}{c|}{0.256$_{+61.0}$} & \multicolumn{2}{c}{\cellcolor[rgb]{ .806,  .802,  .802}0.488$_{+39.0}$} & \multicolumn{2}{c|}{\cellcolor[rgb]{ .806,  .802,  .802}0.726$_{+22.2}$} & \multicolumn{2}{c}{\cellcolor[rgb]{ .806,  .802,  .802}0.416$_{+34.2}$} & \multicolumn{2}{c|}{\cellcolor[rgb]{ .806,  .802,  .802}0.568$_{+30.9}$} & \multicolumn{2}{c}{\cellcolor[rgb]{ .806,  .802,  .802}0.442$_{+26.6}$} & \multicolumn{2}{c}{0.619$_{+22.6}$} \\
    \midrule
    \multicolumn{1}{c}{\multirow{4}[2]{*}{\makecell[c]{SiamMask\\($\sim$14FPS)\\(ResNet50 \cite{He2015ResNet})}}} & N/A   & \multicolumn{2}{c}{0.247$_{+0.00}$} & \multicolumn{2}{c|}{0.313$_{+0.00}$} & \multicolumn{2}{c}{0.455$_{+0.00}$} & \multicolumn{2}{c|}{0.703$_{+0.00}$} & \multicolumn{2}{c}{0.405$_{+0.00}$} & \multicolumn{2}{c|}{0.571$_{+0.00}$} & \multicolumn{2}{c}{0.436$_{+0.00}$} & \multicolumn{2}{c}{0.639$_{+0.00}$} \\
          & $\mathcal{P}_\mathrm{M}$ & \multicolumn{2}{c}{\cellcolor[rgb]{ .806,  .802,  .802}0.370$_{+49.8}$} & \multicolumn{2}{c|}{\cellcolor[rgb]{ .806,  .802,  .802}0.508$_{+62.3}$} & \multicolumn{2}{c}{0.531$_{+16.7}$} & \multicolumn{2}{c|}{0.760$_{+8.10}$} & \multicolumn{2}{c}{0.449$_{+10.9}$} & \multicolumn{2}{c|}{0.607$_{+6.30}$} & \multicolumn{2}{c}{0.532$_{+22.0}$} & \multicolumn{2}{c}{0.743$_{+16.9}$} \\
          & $\mathcal{P}_\mathrm{V}$ & \multicolumn{2}{c}{0.292$_{+18.2}$} & \multicolumn{2}{c|}{0.405$_{+29.4}$} & \multicolumn{2}{c}{0.532$_{+16.9}$} & \multicolumn{2}{c|}{0.777$_{+10.5}$} & \multicolumn{2}{c}{0.430$_{+6.20}$} & \multicolumn{2}{c|}{0.601$_{+5.30}$} & \multicolumn{2}{c}{0.503$_{+15.4}$} & \multicolumn{2}{c}{0.705$_{+10.3}$} \\
          & $\mathcal{P}_\mathrm{MV}$ & \multicolumn{2}{c}{0.342$_{+29.5}$} & \multicolumn{2}{c|}{0.463$_{+47.9}$} & \multicolumn{2}{c}{\cellcolor[rgb]{ .806,  .802,  .802}0.566$_{+24.4}$} & \multicolumn{2}{c|}{\cellcolor[rgb]{ .806,  .802,  .802}0.797$_{+13.4}$} & \multicolumn{2}{c}{\cellcolor[rgb]{ .806,  .802,  .802}0.469$_{+15.8}$} & \multicolumn{2}{c|}{\cellcolor[rgb]{ .806,  .802,  .802}0.644$_{+12.8}$} & \multicolumn{2}{c}{\cellcolor[rgb]{ .806,  .802,  .802}0.536$_{+22.9}$} & \multicolumn{2}{c}{\cellcolor[rgb]{ .806,  .802,  .802}0.749$_{+17.2}$} \\
      \midrule
    \multicolumn{1}{c}{\multirow{4}[2]{*}{\makecell[c]{SiamGAT\\($\sim$16FPS)\\(GoogleNet \cite{szegedy2016GoogleNet})}}} & N/A   & \multicolumn{2}{c}{0.264$_{+0.00}$} & \multicolumn{2}{c|}{0.347$_{+0.00}$} & \multicolumn{2}{c}{0.489$_{+0.00}$} & \multicolumn{2}{c|}{0.720$_{+0.00}$} & \multicolumn{2}{c}{0.475$_{+0.00}$} & \multicolumn{2}{c|}{0.663$_{+0.00}$} & \multicolumn{2}{c}{0.466$_{+0.00}$} & \multicolumn{2}{c}{0.666$_{+0.00}$} \\
          & $\mathcal{P}_\mathrm{M}$ & \multicolumn{2}{c}{0.396$_{+50.0}$} & \multicolumn{2}{c|}{0.520$_{+49.9}$} & \multicolumn{2}{c}{0.549$_{+12.3}$} & \multicolumn{2}{c|}{0.777$_{+7.90}$} & \multicolumn{2}{c}{0.519$_{+9.26}$} & \multicolumn{2}{c|}{0.705$_{+6.33}$} & \multicolumn{2}{c}{0.530$_{+13.7}$} & \multicolumn{2}{c}{\cellcolor[rgb]{ .806,  .802,  .802}0.739$_{+11.0}$} \\
          %TODO 
          & $\mathcal{P}_\mathrm{V}$ & \multicolumn{2}{c}{0.392$_{+48.5}$} & \multicolumn{2}{c|}{0.536$_{+54.4}$} & \multicolumn{2}{c}{0.575$_{+17.6}$} & \multicolumn{2}{c|}{\cellcolor[rgb]{ .806,  .802,  .802}0.801$_{+11.3}$} & \multicolumn{2}{c}{0.503$_{+5.90}$} & \multicolumn{2}{c|}{0.686$_{+3.50}$} & \multicolumn{2}{c}{0.514$_{+10.3}$} & \multicolumn{2}{c}{0.713$_{+7.10}$} \\
          
          & $\mathcal{P}_\mathrm{MV}$ & \multicolumn{2}{c}{\cellcolor[rgb]{ .806,  .802,  .802}0.415$_{+57.2}$} & \multicolumn{2}{c|}{\cellcolor[rgb]{ .806,  .802,  .802}0.561$_{+61.7}$} & \multicolumn{2}{c}{\cellcolor[rgb]{ .806,  .802,  .802}0.583$_{+19.2}$} & \multicolumn{2}{c|}{0.796$_{+10.6}$} & \multicolumn{2}{c}{\cellcolor[rgb]{ .806,  .802,  .802}0.519$_{+9.3}$} & \multicolumn{2}{c|}{\cellcolor[rgb]{ .806,  .802,  .802}0.708$_{+6.80}$} & \multicolumn{2}{c}{\cellcolor[rgb]{ .806,  .802,  .802}0.531$_{+13.9}$} & \multicolumn{2}{c}{0.733$_{+10.1}$} \\
    \bottomrule[1.5pt]
    \end{tabular}%%
    \vspace{-0.4cm}
	\end{threeparttable}
   \label{tab:PVT++}%
\end{table*}%

    For efficiency, our visual predictor $\mathcal{P}_{\mathrm{V}}$ takes search and template features directly from the tracker backbone as input. Besides, we also find the strong representation in the pre-trained tracker models can boost the small-capacity predictor network.  Specifically, template feature $\mathbf{z}\in\mathbb{R}^{1\times C_{\mathrm{V}} \times a \times a}$ is extracted from the given object template patch in the initial frame and search feature $\mathbf{x}_{f_{j}}\in\mathbb{R}^{1\times C_{\mathrm{V}} \times s \times s}$ is obtained from the $f_j$-th frame patch cropped around $(x_{f_{j-1}}, y_{f_{j-1}})$. Given $k$ past search features $\mathbf{x}_{f_{j-k+1}:f_j}\in\mathbb{R}^{k\times C_{\mathrm{V}} \times s \times s}$ and $\mathbf{z}$, we have:
	\begin{equation}
        \small
        \hat{\mathbf{m}}_{f,\mathrm{V}} = \mathcal{P}_\mathrm{V}\big(\mathbf{x}_{f_{j-k+1}:f_j}, \mathbf{z}, \Delta_f\big)\odot{\mathbf{p}_{f_j}}\ .
    \end{equation}
    
    The detailed model structure of $\mathcal{P}_{\mathrm{V}}$ is shown in \fref{fig:structure}(b). Inspired by Siamese trackers~\cite{Li2019SiamRPNEO}, the feature encoding stage adopts $1\times1$ convolution before depth-wise correlation (DW-Corr) to produce the similarity map $\mathbf{x}^{\mathrm{e}}_{f_{j-k+1}:f_j}\in\mathbb{R}^{k\times C_{\mathrm{V}} \times s' \times s'}$. For temporal interaction, we apply 3D convolution and global average pooling. 
	
	We find directly training $\mathcal{P}_{\mathrm{V}}$ meets convergence difficulty (See \sref{sec:6.3}). We hypothesize this is because the intermediate similarity map $\mathbf{x}^{\mathrm{e}}_{f_{j-k+1}:f_j}$ fails to provide explicit motion information. To solve this, we introduce an auxiliary branch $\mathcal{A}$, which takes $\mathbf{x}^{\mathrm{e}}_{f_{j-k+1}:f_j}$ as input to obtain the corresponding motion $\mathbf{m}^{\mathrm{e}}_{f_{j-k+1}:f_j}$,
    \begin{equation}
        \small
        \mathbf{m}^{\mathrm{e}}_{f_{j-k+1}:f_j} = \mathcal{A}(\mathbf{x}^{\mathrm{e}}_{f_{j-k+1}:f_j}).
    \end{equation}
 
    During training, we supervise both the auxiliary branch and the predictive decoder, \textit{i.e.,} $\mathcal{L}_{\mathrm{V}} = \mathcal{L}_1(\hat{\mathbf{m}}_{f,\mathrm{V}}, \mathbf{m}_{f})+\mathcal{L}_1(\mathbf{m}^{\mathrm{e}}_{f_{j-k+1}:f_j}, \mathbf{m}_{f_{j-k+1}:f_j})$.
    
\subsection{Multi-Modality-based Predictor}

    The final predictor $\mathcal{P}_{\mathrm{MV}}$ is constructed as a combination of motion $\mathcal{P}_{\mathrm{M}}$ and visual predictors $\mathcal{P}_{\mathrm{V}}$, which takes both visual feature $\mathbf{x}_{f_{j-k+1}:f_j}, \mathbf{z}$ and motion information $\mathbf{m}_{f_{j-k+1}:f_j}$ as input, \textit{i.e.},    
    \begin{equation}
        \hat{\mathbf{m}}_{f, \mathrm{MV}} = \mathcal{P}_{\mathrm{MV}}\big(\mathbf{m}_{f_{j-k+1}:f_j}, \mathbf{x}_{f_{j-k+1}:f_j}, \mathbf{z}, \Delta_f\big)\odot{\mathbf{p}_{f_j}}\ .
    \end{equation}
    
	As shown in \fref{fig:structure}, the encoding and temporal interaction parts of $\mathcal{P}_{\mathrm{M}}$ and $\mathcal{P}_{\mathrm{V}}$ run in parallel to form the first two stages of $\mathcal{P}_{\mathrm{MV}}$. We concatenate the encoded feature vectors to obtain the multi-modal feature.
	The predictive decoder follows the same structure to obtain future motions $\hat{\mathbf{m}}_{f, \mathrm{MV}}$. We also tried different fusion strategy in \appref{app:earlyfus}.
	
	For training, we add the two additional predictive decoders respectively after motion and visual predictors to help them predict $\hat{\mathbf{m}}_{f, \mathrm{M}}$ and $\hat{\mathbf{m}}_{f, \mathrm{V}}$, which yields the loss $\mathcal{L}_{\mathrm{MV}} = \alpha_{\mathrm{M}}\mathcal{L}_{\mathrm{M}} + \alpha_{\mathrm{V}}\mathcal{L}_{\mathrm{V}} + \mathcal{L}_1(\hat{\mathbf{M}}_{f,\mathrm{MV}}, \mathbf{M}_{f})$. During inference, we only use the joint predictive decoder. 

    \Remark The predictors $\mathcal{P}_{\mathrm{M}}, \mathcal{P}_{\mathrm{V}}$ and $\mathcal{P}_{\mathrm{MV}}$ can be jointly optimized with tracker $\mathcal{T}$, \textit{i.e.}, during training, both input motion and visual feature are from the tracker module with gradient, so that the two modules are deeply coupled.

\section{Experiments}
\label{sec:exp}

\subsection{Implementation Details}

    \myparagraph{Platform and Datasets. }PVT++ is trained on VID \cite{Russakovsky2015VID}, LaSOT \cite{Fan2019LaSOTAH}, and GOT10k \cite{Huang2019GOT10kAL} using one Nvidia A10 GPU. 
    The evaluation takes authoritative UAV tracking datasets, UAV123, UAV20L~\cite{Mueller2016UAV123}, DTB70~\cite{Li2017DTB70}, and UAVDT~\cite{Du2018UAVDT} on typical UAV computing platform, Nvidia Jetson AGX Xavier, for realistic robotic performance. 
    Since the online latency can fluctuate, we run three times and report the average performance. For simplicity, we only consider the tracker's processing latency during evaluation.
	
    \myparagraph{Metrics.} 
	Following \cite{Fu2022GRSM}, we use two basic metrics, the distance precision (DP) based on center location error (CLE) and area under curve (AUC) based on intersection over union. 
	Under e-LAE, different permitted latency $\sigma$ corresponds to different DP and AUC, \textit{i.e.,} DP@La$\sigma$ and AUC@La$\sigma$.
	We use mDP and mAUC to indicate the area under cure for DP@La$\sigma$ and AUC@La$\sigma$, $\sigma\in[0:0.02:1)$.
	
	\myparagraph{Parameters.} 
	For e-LAE, all the evaluated trackers use their official parameters for fairness.
        To represent the most common case, the image frame rate is fixed to $\kappa=30$ frames/s (FPS) in all the online evaluation.
        For the PVT++ models, we use $k=3$ past frames.
        To determine $N$ for different models, we pre-run the trackers 3 times and record the maximum number of skipped frames, so that when the latency of one specific frame fluctuates, PVT++ can always cover the skipped frame and make sufficient predictions.
        Detailed training configurations can be found in \appref{app:config}.

\subsection{Extended Latency-Aware Evaluation}
\label{sec:elae}
	We evaluate a total of 17 SOTA trackers\footnote{Subscripts denote the backbone used, \textit{i.e.,} MobileNet~\cite{sandler2018mobilenetv2}, and ResNet 18 or 50~\cite{He2015ResNet}.} under e-LAE:
	SiamRPN~\cite{Li2018HighPV}, SiamRPN++$_{\mathrm{M}}$~\cite{Li2019SiamRPNEO}, SiamRPN++$_{\mathrm{R}}$~\cite{Li2019SiamRPNEO}, SiamMask~\cite{Wang2019Mask}, SiameseFC++~\cite{Xu2020SiamFC++}, 
	DaSiamRPN~\cite{Zhu2018DistractorawareSN}, SiamAPN~\cite{Cao2021APN}, SiamAPN++~\cite{Cao2021APN++}, HiFT~\cite{Cao2021HiFT}, SiamGAT~\cite{Guo2021SiamGAT},
	SiamBAN~\cite{Chen2020SiamBAN}, SiamCAR~\cite{Guo2020SiamCAR}, ATOM~\cite{Danelljan2019ATOMAT}, DiMP$_{\mathrm{50}}$~\cite{Bhat2019DiMP}, DiMP$_{\mathrm{18}}$~\cite{Bhat2019DiMP}, PrDiMP~\cite{Martin2020PrDiMP}, 
	and TrDiMP~\cite{wang2021TrDiMP}.
	
	As in \fref{fig:elae}, we draw curve plots to reflect their performance in AUC and DP metrics under different permitted latency $\sigma$. We report the [online mAUC and mDP, {\color{gray}offline AUC and DP}] in the legend.
% 	mAUC and mDP (in solid numbers) as well as the offline AUC and DP in the legend. 
	Some \textit{offline} highly accurate trackers like SiamRPN++$_{\mathrm{R}}$ \cite{Li2019SiamRPNEO}, SiamCAR \cite{Guo2020SiamCAR}, SiamBAN \cite{Chen2020SiamBAN}, and ATOM \cite{Danelljan2019ATOMAT} can degrade by up to \textbf{70}\% in our online evaluation setting.
	
    \Remark e-LAE can better assess the real-time trackers. In DTB70, SiamAPN++ and HiFT are two real-time trackers with HiFT more accurate in success. While since SiamAPN++ is faster, its e-LAE performance will be better.

\subsection{Empirical Analysis of PVT++}
\label{sec:6.3}
    % % Att.
    \begin{table}[!t]
        \centering
        \setlength{\tabcolsep}{0.8mm}
        \fontsize{6.5}{7}\selectfont
        \caption{Attribute-based analysis of PVT++ in UAVDT \cite{Du2018UAVDT}. We found different modality has their specific advantage. Together, the joint model can utilize both and is the most robust under complex UAV tracking challenges. \colorbox[rgb]{ .806,  .802,  .802}{Gray} denotes best results.}
        \begin{tabular}{cc|cccccccc}
        \toprule[1.5pt]
              & Metric & \multicolumn{8}{c}{AUC@La0} \\
        \multicolumn{1}{c}{Tracker} & Att.  & BC    & CR    & OR    & SO    & IV    & OB    & SV    & LO \\
        \midrule
        \multirow{4}[1]{*}{SiamRPN++$_{\mathrm{M}}$} & N/A   & 0.448 & 0.450 & 0.438 & 0.494 & 0.539 & 0.525 & 0.490 & 0.422 \\
              & $\mathcal{P}_\mathrm{M}$ & 0.461 & 0.495 & 0.481 & \cellcolor[rgb]{ .806,  .802,  .802}0.549 & 0.578 & 0.542 & 0.505 & \cellcolor[rgb]{ .806,  .802,  .802}0.521 \\
              & $\mathcal{P}_\mathrm{V}$ & 0.504 & 0.520 & 0.538 & 0.525 & 0.588 & 0.568 & 0.584 & 0.436 \\
              & $\mathcal{P}_\mathrm{MV}$ & \cellcolor[rgb]{ .806,  .802,  .802}0.505 & \cellcolor[rgb]{ .806,  .802,  .802}0.535 & \cellcolor[rgb]{ .806,  .802,  .802}0.549 & 0.545 & \cellcolor[rgb]{ .806,  .802,  .802}0.599 & \cellcolor[rgb]{ .806,  .802,  .802}0.589 & \cellcolor[rgb]{ .806,  .802,  .802}0.586 & 0.511 \\
        \midrule
        \multirow{4}[2]{*}{SiamMask} & N/A   & 0.404 & 0.425 & 0.404 & 0.468 & 0.475 & 0.471 & 0.438 & 0.389 \\
              & $\mathcal{P}_\mathrm{M}$ & 0.465 & 0.503 & 0.491 & 0.536 & 0.558 & 0.542 & 0.526 & 0.421 \\
              & $\mathcal{P}_\mathrm{V}$ & 0.488 & 0.498 & 0.504 & 0.495 & 0.563 & 0.527 & 0.541 & 0.494 \\
              & $\mathcal{P}_\mathrm{MV}$ & \cellcolor[rgb]{ .806,  .802,  .802}0.520 & \cellcolor[rgb]{ .806,  .802,  .802}0.522 & \cellcolor[rgb]{ .806,  .802,  .802}0.541 & \cellcolor[rgb]{ .806,  .802,  .802}0.540 & \cellcolor[rgb]{ .806,  .802,  .802}0.596 & \cellcolor[rgb]{ .806,  .802,  .802}0.560 & \cellcolor[rgb]{ .806,  .802,  .802}0.566 & \cellcolor[rgb]{ .806,  .802,  .802}0.520 \\
        \midrule
        \midrule
              & Metric & \multicolumn{8}{c}{DP@La0} \\
        \multicolumn{1}{c}{Tracker} & Att.  & BC    & CR    & OR    & SO    & IV    & OB    & SV    & LO \\
        \midrule
        \multirow{4}[1]{*}{SiamRPN++$_{\mathrm{M}}$} & N/A   & 0.659 & 0.643 & 0.638 & 0.779 & 0.777 & 0.772 & 0.680 & 0.569 \\
              & $\mathcal{P}_\mathrm{M}$ & 0.666 & 0.684 & 0.681 & \cellcolor[rgb]{ .806,  .802,  .802}0.815 & 0.811 & 0.778 & 0.691 & \cellcolor[rgb]{ .806,  .802,  .802}0.717 \\
              & $\mathcal{P}_\mathrm{V}$ & \cellcolor[rgb]{ .806,  .802,  .802}0.733 & 0.720 & 0.753 & 0.793 & 0.835 & 0.822 & \cellcolor[rgb]{ .806,  .802,  .802}0.796 & 0.585 \\
              & $\mathcal{P}_\mathrm{MV}$ & 0.727 & \cellcolor[rgb]{ .806,  .802,  .802}0.732 & \cellcolor[rgb]{ .806,  .802,  .802}0.764 & 0.814 & \cellcolor[rgb]{ .806,  .802,  .802}0.848 & \cellcolor[rgb]{ .806,  .802,  .802}0.846 & 0.794 & 0.694 \\
        \midrule
        \multirow{4}[2]{*}{SiamMask} & N/A   & 0.628 & 0.620 & 0.612 & 0.803 & 0.743 & 0.756 & 0.650 & 0.571 \\
              & $\mathcal{P}_\mathrm{M}$ & 0.672 & 0.702 & 0.709 & 0.818 & 0.797 & 0.802 & 0.729 & 0.590 \\
              & $\mathcal{P}_\mathrm{V}$ & 0.718 & 0.696 & 0.723 & 0.787 & 0.817 & 0.801 & 0.763 & 0.696 \\
              & $\mathcal{P}_\mathrm{MV}$ & \cellcolor[rgb]{ .806,  .802,  .802}0.731 & \cellcolor[rgb]{ .806,  .802,  .802}0.712 & \cellcolor[rgb]{ .806,  .802,  .802}0.752 & \cellcolor[rgb]{ .806,  .802,  .802}0.819 & \cellcolor[rgb]{ .806,  .802,  .802}0.829 & \cellcolor[rgb]{ .806,  .802,  .802}0.813 & \cellcolor[rgb]{ .806,  .802,  .802}0.783 & \cellcolor[rgb]{ .806,  .802,  .802}0.711 \\
        \bottomrule[1.5pt]
        \end{tabular}%
        \vspace{-0.3cm}
    \label{tab:att}%
    \end{table}%
    % Dim ana
    \begin{table}[!t]
        \centering
        \setlength{\tabcolsep}{1.6mm}
            \fontsize{6.5}{7}\selectfont
        \caption{Dimension analysis of different modules in PVT++ on DTB70~\cite{Li2017DTB70} and UAVDT~\cite{Du2018UAVDT}. $\mathrm{Enc.}_{\mathrm{M}}$ and $\mathrm{Enc.}_{\mathrm{V}}$ represent the motion and visual encoders, respectively. $\mathrm{Dec.}_{\mathrm{MV}}$ denotes the joint decoder. * indicates our default setting. We find the channel dimension of PVT++ can be small, so that it introduces very few extra latency on robotics platforms.
        }
        \begin{tabular}{ccc|cccc}
        \toprule[1.5pt]
        \multicolumn{3}{c|}{Dim. of Modules} & \multicolumn{2}{c}{DTB70} & \multicolumn{2}{c}{UAVDT} \\
        $\mathrm{Enc.}_{\mathrm{M}}$ & $\mathrm{Enc.}_{\mathrm{V}}$ & $\mathrm{Dec.}_{\mathrm{MV}}$ & mAUC  & mDP   & mAUC  & mDP \\
        \multicolumn{3}{c|}{Base Tracker} & 0.305 & 0.387 & 0.494 & 0.719 \\
        \midrule
    16    &       &       & 0.357 & 0.479 & 0.565 & 0.797 \\
    32    &       &       & 0.359 & 0.483 & 0.575 & 0.81 \\
    64*   & 64*   & 32*   & \textbf{0.399} & \textbf{0.536} & \textbf{0.576} & \textbf{0.807} \\
    128   &       &       & 0.373 & 0.504 & 0.571 & 0.803 \\
    \midrule
          & 16    &       & 0.362 & 0.487 & 0.545 & 0.772 \\
          & 32    &       & 0.363 & 0.493 & 0.554 & 0.784 \\
    64*   & 64*   & 32*   & \textbf{0.399} & \textbf{0.536} & \textbf{0.576} & \textbf{0.807} \\
          & 128   &       & 0.364 & 0.486 & 0.558 & 0.788 \\
    \midrule
          &       & 16    & 0.369 & 0.496 & 0.572 & 0.804 \\
    64*   & 64*   & 32*   & \textbf{0.399} & \textbf{0.536} & \textbf{0.576} & \textbf{0.807} \\
          &       & 64    & 0.373 & 0.503 & 0.567 & 0.807 \\
          &       & 128   & 0.362 & 0.485 & 0.561 & 0.791 \\
        \bottomrule[1.5pt]
        \end{tabular}%
        \vspace{-0.2cm}
        \label{tab:Dim}%
    \end{table}%

    % ABLA
    \begin{table}[h]
    \centering
    \setlength{\tabcolsep}{0.55mm}
        \fontsize{6.5}{8}\selectfont
    	\caption{Ablation studies on DTB70~\cite{Li2017DTB70}. Official version of PVT++ is marked out in Blackbody. The subscripts $^*$ means predicting raw value instead of motion factor, $^{\dagger}$ denotes training without auxiliary supervision, and $^{\ddagger}$ indicates training with tracker fixed. \textcolor[rgb]{ .753,  0,  0}{Red} denotes improvement and \textcolor[rgb]{ .184,  .459,  .71}{blue} represents dropping.
        }
        \begin{tabular}{cc|cc|ccc|ccc}
        \toprule[1.5pt]
        \multicolumn{2}{c|}{Ablate Module} & \multicolumn{2}{c|}{Motion Factor$^{*}$} & \multicolumn{6}{c}{Auxiliary Supervision$^\dagger$ and Joint Training$^\ddagger$} \\
        \midrule
        Method & Base  & $\mathcal{P}_\mathrm{M}$    & $\mathcal{P}^{*}_\mathrm{M}$   & $\mathcal{P}_\mathrm{V}$    & $\mathcal{P}^\dagger_\mathrm{V}$   & $\mathcal{P}^\ddagger_\mathrm{V}$   & $\mathcal{P}_\mathrm{MV}$   & $\mathcal{P}^\dagger_\mathrm{MV}$  & $\mathcal{P}^\ddagger_\mathrm{MV}$ \\
        AUC@La0 & 0.305 & \textcolor[rgb]{ .753,  0,  0}{\textbf{0.385}} & \textcolor[rgb]{ .184,  .459,  .71}{0.300} & \textcolor[rgb]{ .753,  0,  0}{\textbf{0.352}} & \textcolor[rgb]{ .753,  0,  0}{0.311} & \textcolor[rgb]{ .184,  .459,  .71}{0.278} & \textcolor[rgb]{ .753,  0,  0}{\textbf{0.399}} & \textcolor[rgb]{ .753,  0,  0}{0.323} & \textcolor[rgb]{ .184,  .459,  .71}{0.294} \\
        Delta\% & 0.00  & \textcolor[rgb]{ .753,  0,  0}{\textbf{26.2}} & \textcolor[rgb]{ .184,  .459,  .71}{-1.60} & \textcolor[rgb]{ .753,  0,  0}{\textbf{15.4}} & \textcolor[rgb]{ .753,  0,  0}{2.00} & \textcolor[rgb]{ .184,  .459,  .71}{-8.90} & \textcolor[rgb]{ .753,  0,  0}{\textbf{30.8}} & \textcolor[rgb]{ .753,  0,  0}{5.90} & \textcolor[rgb]{ .184,  .459,  .71}{-3.60} \\
        DP@La0 & 0.387 & \textcolor[rgb]{ .753,  0,  0}{\textbf{0.523}} & \textcolor[rgb]{ .184,  .459,  .71}{0.383} & \textcolor[rgb]{ .753,  0,  0}{\textbf{0.472}} & \textcolor[rgb]{ .753,  0,  0}{0.412} & \textcolor[rgb]{ .184,  .459,  .71}{0.349} & \textcolor[rgb]{ .753,  0,  0}{\textbf{0.536}} & \textcolor[rgb]{ .184,  .459,  .71}{0.429} & \textcolor[rgb]{ .184,  .459,  .71}{0.387} \\
        Delta\% & 0.00  & \textcolor[rgb]{ .753,  0,  0}{\textbf{35.1}} & \textcolor[rgb]{ .184,  .459,  .71}{-1.00} & \textcolor[rgb]{ .753,  0,  0}{\textbf{22.0}} & \textcolor[rgb]{ .753,  0,  0}{6.50} & \textcolor[rgb]{ .184,  .459,  .71}{-9.80} & \textcolor[rgb]{ .753,  0,  0}{\textbf{38.5}} & \textcolor[rgb]{ .184,  .459,  .71}{-10.9} & \textcolor[rgb]{ .184,  .459,  .71}{0.00} \\
        \bottomrule[1.5pt]
        \end{tabular}%
        \vspace{-0.3cm}
      \label{tab:abla}%
    \end{table}%

    \myparagraph{Overall Effect.} To evaluate PVT++, we construct predictive trackers with four well-known methods, \textit{i.e.,} SiamRPN++$_\mathrm{M}$~\cite{Li2019SiamRPNEO}, SiamRPN++$_\mathrm{R}$~\cite{Li2019SiamRPNEO}, SiamMask~\cite{Wang2019Mask}, and SiamGAT~\cite{Guo2021SiamGAT}.
    As in \tref{tab:PVT++}, with PVT++, their online performance can be significantly boosted by up to \textbf{60}\%, which sometimes is better than their \textit{offline} performance. PVT++ also works for recent transformer-based trackers \cite{ye2022joint,cui2022mixformer}, the results can be found in \appref{app:tf}.
    
    \Remark Real-time trackers~\cite{Cao2021APN,Cao2021APN++,Cao2021HiFT,Li2018HighPV} perform generally better than non-real-time ones in online evaluation. While we observe that non-real-time trackers empowered by PVT++ can notably outperform real-time ones. \textit{E.g.}, SiamRPN++$_\mathrm{M}$~\cite{Li2019SiamRPNEO} with $P_{\mathrm{MV}}$ achieves an amazing \textbf{0.807} mDP in UAVDT, better than SiamFC++~\cite{Xu2020SiamFC++} (0.761).
	
    \myparagraph{Attribute-based Analysis. }
    For a comprehensive evaluation, we follow \cite{Du2018UAVDT} and evaluate PVT++ on various challenge attributes\footnote{Background cluter (BC), camera rotation (CR), object rotation (OR), small object (SO), illumination variation (IV), object blur (OB), scale variation (SV), and large occlusion (LO).}. 
    From \tref{tab:att}, We found that motion and vision have advantages in different attributes. 
    $\mathcal{P}_{\mathrm{V}}$ improves CR and OR, while $\mathcal{P}_{\mathrm{M}}$ is good at SO and LO. The joint model $\mathcal{P}_{\mathrm{MV}}$ makes use of both and is the most robust under various complex aerial tracking challenges. For the full attribute analysis, please see \appref{app:att}.

     \myparagraph{Dimension Analysis. }In addition to its promising performance, PVT++ can also work with very small capacity, which contributes to its lightweight architecture and high efficiency on low-powered UAVs. 
     We analyse the modules of PVT++ with different feature channels in \tref{tab:Dim}, where $64$ channels for encoders ($\mathrm{Enc.}_{\mathrm{M}}$, $\mathrm{Enc.}_{\mathrm{V}}$) and $32$ channels for the joint decoder ($\mathrm{Dec.}_{\mathrm{J}}$) work best.
     We present more efficiency and complexity comparisons with other motion predictors \cite{zhang2019srlstm,gupta2018social,liang2019next} in \appref{app:eff}

    \myparagraph{Ablation Studies. }We ablate the effect of motion factor prediction, auxiliary branch, and the joint training of PVT++ on DTB70~\cite{Li2017DTB70} with SiamRPN++$_\mathrm{M}$ in \tref{tab:abla}. 
    Compared with directly predicting the motion value ($\mathcal{P}^\dagger_\mathrm{M}$), using \textit{motion factor} as the prediction target ($\mathcal{P}_\mathrm{M}$) can yield much better performance. 
    Removing \textit{auxiliary branch} $\mathcal{A}$ in $\mathcal{P}_{\mathrm{V}}$ and $\mathcal{P}_{\mathrm{MV}}$ to be $\mathcal{P}^\dagger_{\mathrm{V}}$ and $\mathcal{P}^\dagger_{\mathrm{MV}}$, we observe a significant performance drop due to the difficulty in convergence. 
    \textit{Joint training} the tracker and the predictor ($\mathcal{P}_{\mathrm{V}}$ \& $\mathcal{P}_{\mathrm{MV}}$) perform much better than fixing the tracker ($\mathcal{P}^\ddagger_{\mathrm{V}}$ and $\mathcal{P}^\ddagger_{\mathrm{MV}}$). Training loss of the ablation studies are visualized in \appref{app:training_visual}.

    \subsection{Comparison with KF-based Solutions}

    % KF vs PVT
    % Table generated by Excel2LaTeX from sheet 'Tab2'
    \begin{table}[!t]
        \centering
        \setlength{\tabcolsep}{0.6mm}
        \fontsize{5.9}{6.1}\selectfont
        \caption{Averaged results comparison on four datasets \cite{Mueller2016UAV123,Du2018UAVDT,Li2017DTB70}. The motion based PVT++ can achieve on par or better results than prior KF-based solutions \cite{Li2020TowardsSP,Li2020PredictiveVT}. Further introducing visual cues, PVT++ can acquire higher robustness. KF$^\dagger$ and $^\ddagger$ denotes learnable baselines \cite{piga2021differentiable,Haarnoja2016Backprop}, which are still less robust than PVT++.}
        \begin{tabular}{cc|ccc|ccccc}
        \toprule[1.5pt]
        \multicolumn{2}{c|}{Type} & \multicolumn{3}{c|}{Tradition Model} & \multicolumn{5}{c}{Learning-based} \\
        Tracker & Pred. & N/A   & KF\cite{Li2020TowardsSP}    & PVT\cite{Li2020PredictiveVT}   & KF$^\dagger$ \cite{piga2021differentiable} & KF$^\ddagger$ \cite{Haarnoja2016Backprop} & $\mathcal{P}_\mathrm{M}$  & $\mathcal{P}_\mathrm{V}$  & $\mathcal{P}_\mathrm{MV}$ \\
        \midrule
        \multirow{2}[2]{*}{SiamRPN++$_{\mathrm{M}}$} & AUC@La0 & 0.43  & 0.462 & 0.473 & 0.466 & 0.481 & 0.483 & 0.477 & \textcolor[rgb]{ .753,  0,  0}{\textbf{0.505}} \\
              & DP@La0 & 0.601 & 0.639 & 0.651 & 0.642 & 0.658 & 0.663 & 0.662 & \textcolor[rgb]{ .753,  0,  0}{\textbf{0.695}} \\
        \midrule
        \midrule
        \multirow{2}[2]{*}{SiamMask} & AUC@La0 & 0.386 & 0.441 & 0.465 & 0.458 & 0.468 & 0.471 & 0.439 & \textcolor[rgb]{ .753,  0,  0}{\textbf{0.478}} \\
              & DP@La0 & 0.557 & 0.607 & 0.639 & 0.631 & 0.638 & 0.655 & 0.622 & \textcolor[rgb]{ .753,  0,  0}{\textbf{0.663}} \\
        \midrule
        \midrule
        \multirow{2}[2]{*}{SiamRPN++$_{\mathrm{R}}$} & AUC@La0 & 0.287 & 0.361 & 0.374 & 0.376 & 0.386 & 0.374 & 0.345 & \textcolor[rgb]{ .753,  0,  0}{\textbf{0.388}} \\
              & DP@La0 & 0.423 & 0.502 & 0.523 & 0.527 & 0.532 & 0.532 & 0.499 & \textcolor[rgb]{ .753,  0,  0}{\textbf{0.542}} \\
        \bottomrule[1.5pt]
        \end{tabular}%
      \label{tab:kf}%
      \vspace{-0.2cm}
    \end{table}%

        % vis_kf
    \begin{figure}[!t]
        \centering
        \includegraphics[width=0.9\columnwidth]{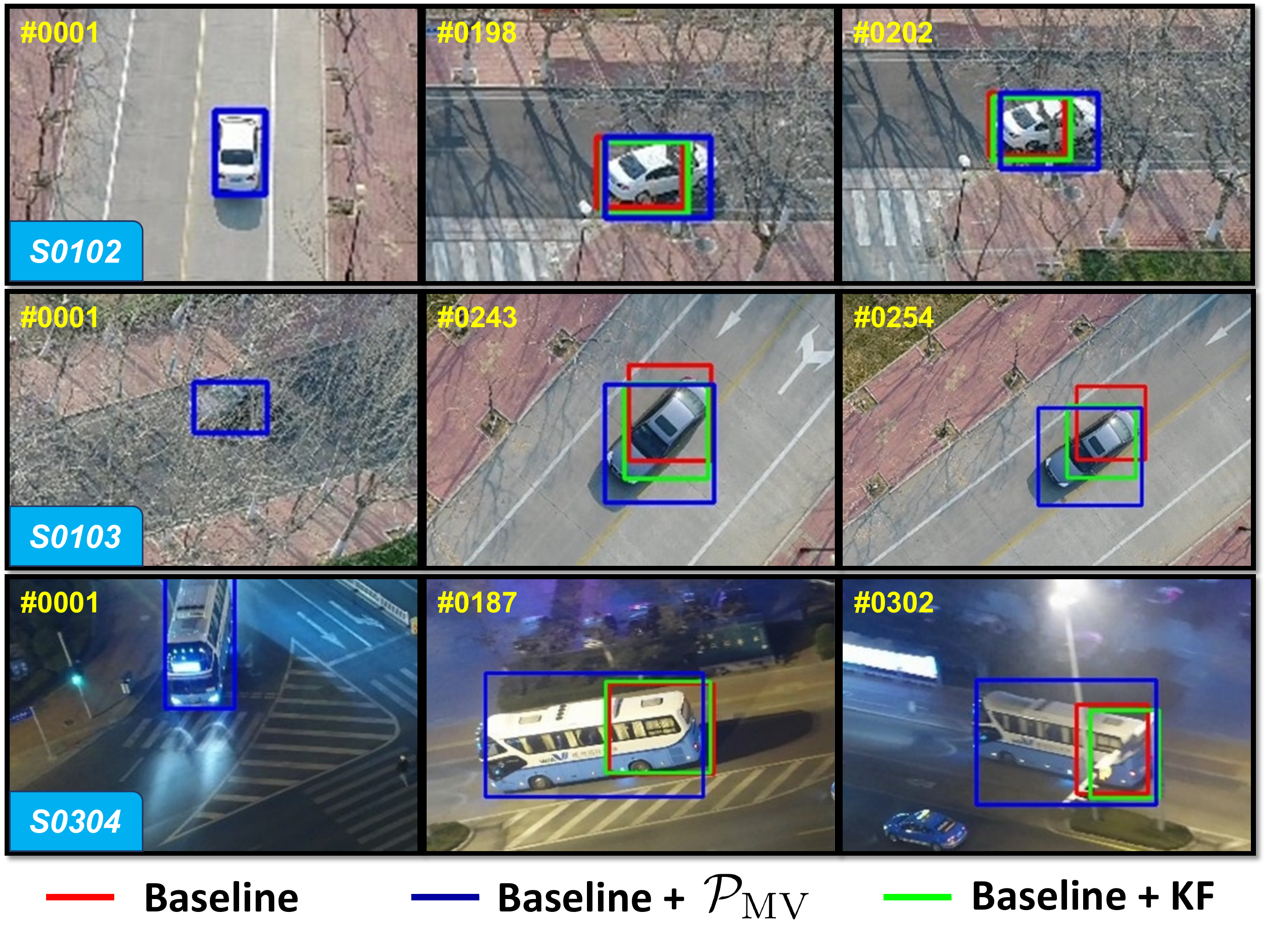}
        \vspace{-0.3cm}
        \caption{Prediction comparison from UAVDT~\cite{Du2018UAVDT}. We use \textcolor{red}{red} lines to demonstrate the original trackers, \textcolor{green}{green} for the KF~\cite{Li2020TowardsSP} prediction, and \textcolor{blue}{blue} for PVT++ prediction. Compared to KF, PVT++ is better at handling challenges like rotation, scale variation, and view point change. Best viewed in color.}
        \vspace{-0.4cm}
        \label{fig:viskf}
    \end{figure}

    \myparagraph{Quantitative Results. }
        Prior attempts to latency-aware perception \cite{Li2020TowardsSP,Li2020PredictiveVT} have introduced model-based approach, \textit{i.e.}, KF \cite{kalman1960}, as predictors. 
        Based on traditional KF, we also designed stronger learnable baselines, KF$^\dagger$ \cite{piga2021differentiable} and KF$^\ddagger$ \cite{Haarnoja2016Backprop}, which adopt the same training as PVT++ models.
        Basically, KF$^\dagger$ \cite{piga2021differentiable} learns the two noise matrix and KF$^\ddagger$ denotes joint training of KF$^\dagger$ and trackers via backpropagation \cite{Haarnoja2016Backprop}.
        We compare these KF-based solutions with PVT++ in \tref{tab:kf}, where the same base tracker models are adopted. 
        We present averaged mAUC and mDP in 4 datasets, DTB70 \cite{Li2017DTB70}, UAVDT~\cite{Du2018UAVDT}, UAV20L~\cite{Mueller2016UAV123}, and UAV123~\cite{Mueller2016UAV123}.
        Compared with the KFs \cite{Li2020TowardsSP,Li2020PredictiveVT}, our learning framework holds the obvious advantage in complex UAV tracking scenes. 
        We also observed that PVT++ is very efficient and introduces very little extra latency on the trackers.
        For specific results per dataset, please refer to \appref{app:allkf}.
     
     \myparagraph{Qualitative Results. }
        To better present the priority of PVT++, some representative scenes are displayed in \fref{fig:viskf}. Given the same history trajectory, PVT++ holds its advantage against KF-based solution~\cite{Li2020TowardsSP}. Especially, when UAV tracking challenges like in-plane rotation (sequence \textit{S0103}) and aspect ration change (sequence \textit{S0304}) appear, PVT++ is capable of fully utilizing the appearance change for robust prediction while simple motion-based KF easily fails.

    \Remark Different from the KFs, which predict future results one by one, PVT++ outputs $N$ results in a single forward pass, resulting in its high efficiency (especially for the motion predictor).

    Apart from the robustness priority, PVT++ is also easier to be deployed. Once trained, no further tuning is needed for PVT++ to fit various scenes. Differently, the noise matrix of the KFs is dependent on the environment, which is hard to tune and may not generalize well to various complex UAV tracking scenes.
	
\subsection{Real-World Tests}
    
    % Real-world test
    \begin{figure}[!t]
    	\centering
    	\includegraphics[width=1\columnwidth]{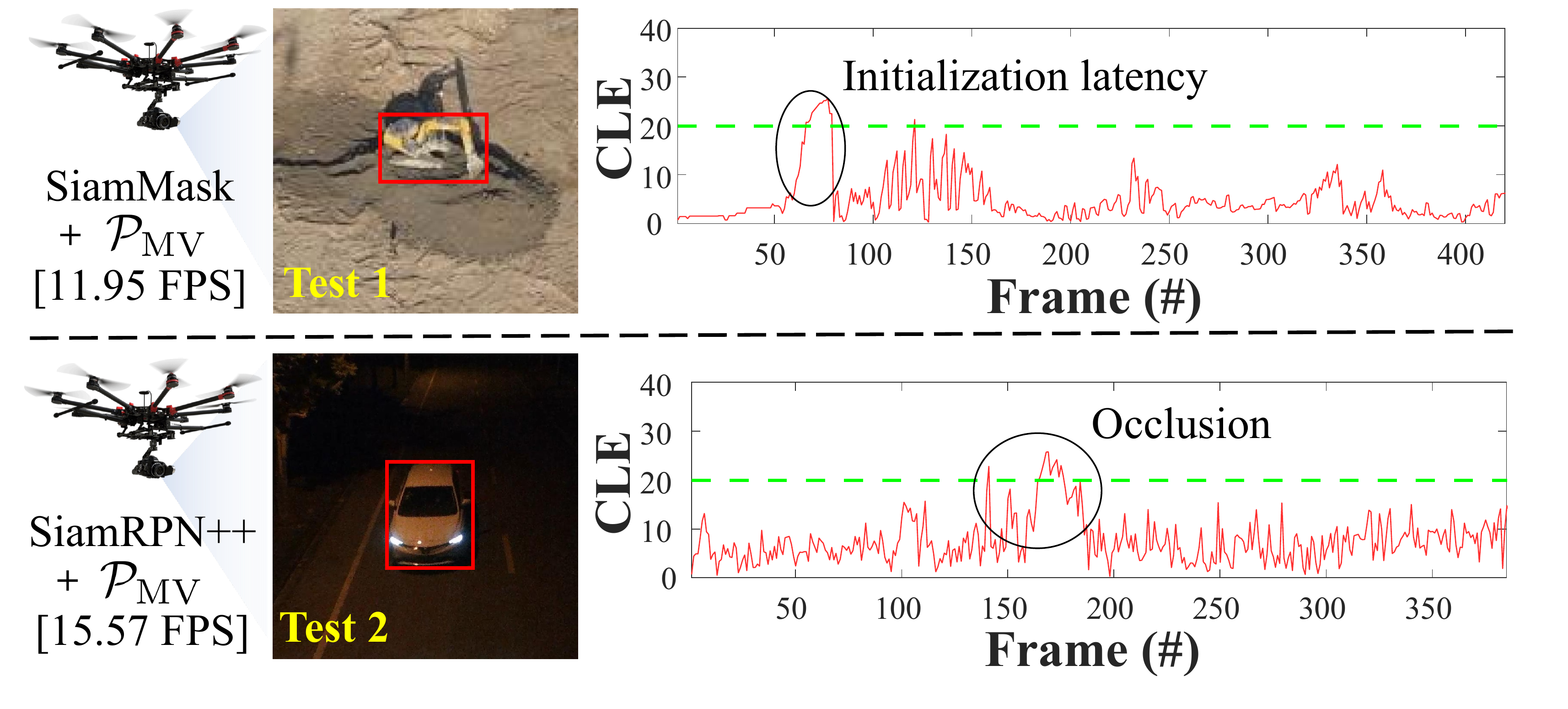}
     \vspace{-0.7cm}
    	\caption{Real-world tests of PVT++. Thanks to PVT++, the non-real-time trackers work effectively under real-world challenges like object deformation in Test 1 and occlusion in Test 2.}
     \vspace{-0.4cm}
    	\label{fig:real}
    \end{figure}

    We further deploy SiamMask \cite{Wang2019Mask} ($\sim$11FPS) and SiamRPN++$_{\mathrm{M}}$ \cite{Li2019SiamRPNEO} ($\sim$15FPS) with PVT++ on a UAV with Nvidia Jetson AGX Xavier as onboard processor. The onboard speed and center location error (CLE) results are shown in \fref{fig:real}. Despite that the original tracker is not real-time, our PVT++ framework can convert it into a predictive tracker and achieve a good result (CLE $<$ 20 pixels) in real-world tracking. More tests see \appref{app:real_world} and the video.

\section{Conclusion}
\label{sec:conclusion}
% \todo{I think you need to discuss the speed, prediction, computation tradeoff a bit more here}
    % \textbf{Limitation. }(1) Consider SiamRPN++$_\mathrm{M}$ with $\sim45$ms latency, PVT++ introduces an extra latency of $\sim5$ms per frame, which is higher than KF ($2\sim3$ms), slightly affecting the performance (further discussion in \appref{app:extra_latency}).
    % (2) For e-LAE, the performance is usually influenced by the state of the hardware, which requires multiple runs for a proper assessment. 
    
    In this work, we present a simple end-to-end framework for latency-aware visual tracking, PVT++, which largely compensates for onboard latency. 
    PVT++ integrates a lightweight predictor module that discovers the visual representation from pre-trained trackers for precise and robust future state estimation.
    To bridge the training-evaluation domain gap, we propose the relative motion factor, which yields a generalizable framework. 
    In addition to PVT++, we introduce extended latency-aware evaluation benchmark (e-LAE), which assesses an \textit{any-speed} tracker in the \textit{online} setting.
    Extensive evaluations on robotics platform from the challenging aerial perspective show the effectiveness of our PVT++, which improves the \textit{offline} tracker by up to \textbf{60}\% in the \textit{online} setting. Real-world tests are further conducted to exhibit the efficacy of PVT++ on physical robots.
    % We hope that PVT++ can facilitate more research on latency-aware visual tracking for real-world robotic applications.

% %%%%%%%%% REFERENCES
{\small
\bibliographystyle{ieee_fullname}
\bibliography{egbib}
}

\newpage
\clearpage

\appendix
\section*{Appendix}
\renewcommand{\thetable}{\Roman{table}}
\renewcommand{\thefigure}{\Roman{figure}}
\renewcommand\thesection{\Alph {section}}
\setcounter{section}{0}
\setcounter{figure}{0}
\setcounter{table}{0}

\section{Overview}
% \subsection*{Overview}
To make our end-to-end predictive visual tracking framework (PVT++) reproducible, we present the detailed configuration in \appref{app:config}, covering the specific model structure, the training settings (with specific hyper-parameters), and the inference settings. Moreover, we provide the PVT++ code library and official models to ensure reproducability.
For clear reference of the notations used in method section, we provide a notation table in \appref{app:notation}.
In \appref{app:visual}, we display representative qualitative visualization results from the authoritative datasets, UAV123~\cite{Mueller2016UAV123}, UAV20L~\cite{Mueller2016UAV123}, DTB70~\cite{Li2017DTB70}, and UAVDT~\cite{Du2018UAVDT}, where the superiority of our PVT++ is clearly shown.
In \appref{app:allkf}, we present detailed results comparison between KF~\cite{kalman1960} and PVT++ to better demonstrate the superiority of our method.
In addition to convolution neural network backbone\cite{He2015ResNet,sandler2018mobilenetv2,szegedy2016GoogleNet}-based trackers, PVT++ further works for transformer-based ones \cite{ye2022joint,cui2022mixformer}, which is presented in \appref{app:tf}.
In \appref{app:eff}, we show that PVT++ is more efficiency and introduces much less extra latency onboard compared with other trajectory preductors~\cite{gupta2018social,zhang2019srlstm,liang2019next}.
We also tried to fuse the motion and visual cues earlier in \appref{app:earlyfus}, where we give an analysis to the strategy adopted in PVT++.
The full attribute-based results from all the four datasets~\cite{Mueller2016UAV123,Li2017DTB70,Du2018UAVDT} are reported in \appref{app:att}, where we exhaustively analyse the specific advantages of two modalities for prediction under various UAV tracking challenges.
The training process of different PVT++ models is visualized in \appref{app:training_visual}, where we present the loss curves to indicate the converging process.
The extra latency introduced by the PVT++ predictor modules is unavoidable, which can have some negative effect to online performance. We provide such analysis in \appref{app:extra_latency}.
We further find PVT++ is capable of converging well in smaller training set (using only 3563 videos from Imagenet VID~\cite{Russakovsky2015VID}), which is shown in \appref{app:training_set}.
Finally, we present additional real-world tests in \appref{app:real_world}, covering more target objects and tracking scenes.

\section{Detailed Configuration}\label{app:config}

\begin{table}[!t]
    \centering
    \setlength{\tabcolsep}{0.4mm}
    \fontsize{7}{8}\selectfont
    \caption{Detailed structure and output sizes of PVT++ models. We use subscript to distinguish between different layers. The output sizes correspond to $B$ batch input.}
    \begin{threeparttable}
    \begin{tabular}{c|ccccc}
    \toprule[1.5pt]
    Branch & Layer & Kernel & $C_{\mathrm{in}}$ & $C_{\mathrm{out}}$ & Out. Size \\
    \midrule
    \multirow{3}[0]{*}{Motion} & FC    & -     & 8     & 32    & $B\times k\times32$ \\
          & 1D Conv & 3     & 32    & 32    & $B\times k \times 32$ \\
          & Avg. Pool & -     & 32    & 32    & $B\times32$ \\
    \midrule
    \multirow{8}[0]{*}{Visual} & 2D Conv$_{\mathrm{t}}$ & $3\times3$ & 256   & 64    & $B\times k\times64\times29\times29$ \\
          & 2D Conv$_{\mathrm{s}}$ & $3\times3$ & 256   & 64    & $B\times k\times64\times25\times25$ \\
          & 2D Conv$_{\mathrm{e}}$ & $1\times1$ & 64    & 64    & $B\times k\times64\times25\times25$ \\
          & 3D Conv & $3\times3\times3$ & 64    & 64    & $B\times k\times64\times25\times25$ \\
          & Avg. Pool & -     & 64    & 64    & $B\times 64$ \\
          & 2D Conv$_{\mathrm{a}}$ & $1\times1$ & 64    & 64    & $B\times k\times64\times25\times25$ \\
          & 2D Conv$_{\mathrm{a}}$ & $1\times1$ & 64    & 4     & $B\times k\times4\times25\times25$ \\
          & Avg. Pool$_{\mathrm{a}}$ & -     & 4     & 4     & $B\times k\times4$ \\
    \midrule
    \multirow{3}[1]{*}{Shared} & FC    & -     & [32, 64, 96] & 32    & $B\times 32$ \\
          & FC    & -     & 32    & 32    & $B\times N\times 32$ \\
          & FC    & -     & 32    & 4     & $B\times N\times 4$ \\
    \bottomrule[1.5pt]
    \end{tabular}%
    \end{threeparttable}
    \label{tab:detai_model}
\end{table}%

\begin{table}[!t]
    \centering
    \setlength{\tabcolsep}{0.7mm}
    \fontsize{5.5}{8}\selectfont
    \caption{List of the important notations in this work.}
    \begin{tabular}{ccc}
        \toprule[1.5pt]
        \textbf{Symbol}	& \textbf{Meaning}  & \textbf{Dimension}  \\
        \midrule
        $f$ & World frame number & $\mathbb{R}$\\
        $\mathcal{I}_f$ & $f$-th image frame & $\mathbb{R}^{W\times H\times3}$ \\
        $j$ & Serial number of the processed frame  & $\mathbb{R}$\\
        $f_j$ & World frame id of the processed $j$-th frame & $\mathbb{R}$\\
        $t^{\mathrm{W}}_f$ & World timestamp & $\mathbb{R}$\\
        $t^{\mathrm{T}}_{f_j}$ & Tracker timestamp & $\mathbb{R}$\\
        $\phi{(f)}, \phi{(f)}_e $ & Input frame id to be paired with frame $f$ & $\mathbb{R}$\\
        $\sigma$ & Permitted latency during evaluation & $\mathbb{R}$\\
        $\mathbf{r}_{f} = [x_f, y_f, w_f, h_f]$ & Raw output by the tracker in frame $f$ & $\mathbb{R}^{1\times4}$\\
        $\hat{\mathbf{b}}_{f} = [\hat{x}_f, \hat{y}_f, \hat{w}_f, \hat{h}_f]$ & Final output bounding box to be evaluated & $\mathbb{R}^{1\times4}$\\
        
        $\mathcal{T}$ & Tracker model & $-$\\
        $\mathcal{P}$ & Predictor model & $-$\\
        
        $\mathbf{m}_{f_{j}}$ & Normalized input motion from $f_{j-1}$ to $f_j$ & $\mathbb{R}^{1\times4}$\\
        $\mathbf{p}_{f_j}$ & Average moving speed from $f_{j-k+1}$ to $f_j$ & $\mathbb{R}^{1\times4}$\\
        $\hat{\mathbf{m}}_{f}$ & Predicted motion from $\phi(f)$ to $f$ & $\mathbb{R}^{1\times4}$\\
        $\mathbf{m}_{f}$ & Ground-truth motion from $\phi(f)$ to $f$ & $\mathbb{R}^{1\times4}$\\
        $\Delta_f$ & Frame interval between the latest and the $f$-th frame & $\mathbb{R}$\\
        $\Delta_{\hat{x}}(f), \Delta_{\hat{y}}(f)$ & Predicted distance between the $f$-th and $\phi(f)$-th frame & $\mathbb{R}$\\
        $\Delta_x({f_{j}}), \Delta_y({f_{j}})$ & Distance from $\mathbf{r}_{f_j}$ to $\mathbf{r}_{f_{j-1}}$ & $\mathbb{R}$\\
    
        $\mathbf{x}_{\phi{(f)}}$ & Search patch feature in frame $\phi(f)$& $\mathbb{R}^{C\times W\times H}$ \\
        $\mathbf{z}$ & Template feature & $\mathbb{R}^{C\times a\times a}$ \\
        $k (=3)$ & Number of past frames& $\mathbb{R}$ \\
        $N$ & Number of the parallel FC layers in the decoder & $\mathbb{R}$ \\
        \bottomrule[1.5pt]
    \end{tabular}%
    \label{tab:notation}%
    \vspace{-0.3cm}
\end{table}%

\begin{table*}[h]
\centering
\setlength{\tabcolsep}{1.2mm}
\fontsize{8}{9}\selectfont
\caption{Per dataset results of different predictor modules. For all the three base trackers in various datasets, our PVT++ generally outperforms previous standard KF solutions~\cite{Li2020TowardsSP,Li2020PredictiveVT} and stronger learnable KF baselines, KF$^{\dagger}$ and KF$^{\ddagger}$.}
    \begin{threeparttable}
        \begin{tabular}{cc|cc|cc|cc|cc}
        \toprule[1.5pt]
        \multicolumn{2}{c|}{Dataset} & \multicolumn{2}{c|}{DTB70} & \multicolumn{2}{c|}{UAVDT} & \multicolumn{2}{c|}{UAV20L} & \multicolumn{2}{c}{UAV123} \\
        Tracker & Pred. & AUC@La0 & DP@La0 & AUC@La0 & DP@La0 & AUC@La0 & DP@La0 & AUC@La0 & DP@La0 \\
        \midrule
        \multicolumn{1}{c}{\multirow{8}[2]{*}{\makecell[c]{SiamRPN++$_{\mathrm{M}}$\\(21FPS)}}} & N/A   & 0.305 & 0.387 & 0.494 & 0.719 & 0.448 & 0.619 & 0.472 & 0.678 \\
        & KF \cite{Li2020TowardsSP}    & 0.349 & 0.482 & 0.527 & 0.737 & 0.458 & 0.624 & 0.515 & 0.712 \\
        & PVT \cite{Li2020PredictiveVT}   & 0.377 & 0.518 & 0.533 & 0.740 & 0.458 & 0.624 & 0.522 & 0.722 \\
        & KF$^\dagger$ \cite{piga2021differentiable}   & 0.367 & 0.504 & 0.519 & 0.732 & 0.466 & 0.630 & 0.511 & 0.703 \\
        & KF$^\ddagger$ \cite{Haarnoja2016Backprop}  & 0.365 & 0.496 & 0.563 & 0.780 & 0.483 & 0.658 & 0.513 & 0.598 \\
        & $\mathcal{P}_\mathrm{M}$ (\textbf{Ours}) & 0.385 & 0.523 & 0.529 & 0.745 & 0.481 & 0.647 & 0.537 & 0.737 \\
        & $\mathcal{P}_\mathrm{V}$ (\textbf{Ours}) & 0.352 & 0.472 & 0.564 & 0.799 & 0.488 & 0.675 & 0.504 & 0.703 \\
        & $\mathcal{P}_\mathrm{MV}$ (\textbf{Ours}) & \textcolor[rgb]{ .753,  0,  0}{\textbf{0.399}} & \textcolor[rgb]{ .753,  0,  0}{\textbf{0.536}} & \textcolor[rgb]{ .753,  0,  0}{\textbf{0.576}} & \textcolor[rgb]{ .753,  0,  0}{\textbf{0.807}} & \textcolor[rgb]{ .753,  0,  0}{\textbf{0.508}} & \textcolor[rgb]{ .753,  0,  0}{\textbf{0.697}} & \textcolor[rgb]{ .753,  0,  0}{\textbf{0.537}} & \textcolor[rgb]{ .753,  0,  0}{\textbf{0.741}} \\
        \midrule
        \midrule
        \multicolumn{1}{c}{\multirow{8}[2]{*}{\makecell[c]{SiamMask\\(12FPS)}}} & N/A   & 0.247 & 0.313 & 0.455 & 0.703 & 0.405 & 0.571 & 0.436 & 0.639 \\
        & KF \cite{Li2020TowardsSP}   & 0.294 & 0.407 & 0.535 & 0.758 & 0.436 & 0.582 & 0.499 & 0.679 \\
        & PVT \cite{Li2020PredictiveVT}  & 0.362 & 0.504 & 0.539 & 0.751 & 0.443 & 0.598 & 0.514 & 0.701 \\
        & KF$^\dagger$ \cite{piga2021differentiable}   & 0.349 & 0.486 & 0.530 & 0.749 & 0.440 & 0.588 & 0.513 & 0.702 \\
        & KF$^\ddagger$ \cite{Haarnoja2016Backprop}  & 0.348 & 0.468 & 0.558 & 0.775 & 0.465 & 0.629 & 0.502 & 0.683 \\
        & $\mathcal{P}_\mathrm{M}$ (\textbf{Ours}) & \textcolor[rgb]{ .753,  0,  0}{\textbf{0.370}} & \textcolor[rgb]{ .753,  0,  0}{\textbf{0.508}} & 0.531 & 0.760 & 0.449 & 0.607 & 0.532 & 0.743 \\
        & $\mathcal{P}_\mathrm{V}$ (\textbf{Ours}) & 0.292 & 0.405 & 0.532 & 0.777 & 0.430 & 0.601 & 0.503 & 0.705 \\
        & $\mathcal{P}_\mathrm{MV}$ (\textbf{Ours}) & 0.342 & 0.463 & \textcolor[rgb]{ .753,  0,  0}{\textbf{0.566}} & \textcolor[rgb]{ .753,  0,  0}{\textbf{0.797}} & \textcolor[rgb]{ .753,  0,  0}{\textbf{0.469}} & \textcolor[rgb]{ .753,  0,  0}{\textbf{0.644}} & \textcolor[rgb]{ .753,  0,  0}{\textbf{0.536}} & \textcolor[rgb]{ .753,  0,  0}{\textbf{0.749}} \\
        \midrule
        \midrule
        \multicolumn{1}{c}{\multirow{8}[2]{*}{\makecell[c]{SiamRPN++$_{\mathrm{M}}$\\(21FPS)}}} & N/A   & 0.136 & 0.159 & 0.351 & 0.594 & 0.310 & 0.434 & 0.349 & 0.505 \\
        & KF \cite{Li2020TowardsSP}    & 0.189 & 0.232 & 0.451 & 0.667 & 0.387 & 0.528 & 0.415 & 0.582 \\
        & PVT \cite{Li2020PredictiveVT}   & 0.201 & 0.254 & 0.467 & 0.687 & 0.396 & 0.547 & 0.434 & 0.605 \\
        & KF$^\dagger$ \cite{piga2021differentiable}  & 0.200 & 0.254 & 0.460 & 0.680 & 0.412 & \textcolor[rgb]{ .753,  0,  0}{0.572} & 0.433 & 0.603 \\
        & KF$^\ddagger$ \cite{Haarnoja2016Backprop} & 0.204 & 0.252 & 0.504 & 0.728 & 0.406 & 0.549 & 0.432 & 0.599 \\
        & $\mathcal{P}_\mathrm{M}$ (\textbf{Ours}) & 0.199 & \textcolor[rgb]{ .753,  0,  0}{\textbf{0.258}} & 0.449 & 0.684 & 0.404 & 0.560 & \textcolor[rgb]{ .753,  0,  0}{\textbf{0.442}} & \textcolor[rgb]{ .753,  0,  0}{\textbf{0.627}} \\
        & $\mathcal{P}_\mathrm{V}$ (\textbf{Ours}) & 0.179 & 0.225 & 0.403 & 0.665 & 0.398 & 0.548 & 0.398 & 0.559 \\
        & $\mathcal{P}_\mathrm{MV}$ (\textbf{Ours}) & \textcolor[rgb]{ .753,  0,  0}{\textbf{0.205}} & 0.256 & \textcolor[rgb]{ .753,  0,  0}{\textbf{0.488}} & \textcolor[rgb]{ .753,  0,  0}{\textbf{0.726}} & \textcolor[rgb]{ .753,  0,  0}{\textbf{0.416}} & 0.568 & 0.442 & 0.619 \\
        \bottomrule[1.5pt]
        \end{tabular}%
        \end{threeparttable}
    \label{tab:full_kf}%
\end{table*}%

    \myparagraph{Specific Model Structure.} Corresponding to Fig. 4 in the paper, we present the detailed model structure of each layer in \tref{tab:detai_model}. Consider $B$ batch inputs and $k$ history frames, the output sizes are also shown in \tref{tab:detai_model} for clear reference. Subscripts are used to distinguish between different layers, \textit{i.e.}, $\cdot_{\mathrm{t}}$ denotes encoding layer for template feature, $\cdot_{\mathrm{s}}$ denotes encoding layer for search feature, $\cdot_{\mathrm{e}}$ denotes encoding layer for the similarity map. $\cdot_{\mathrm{a}}$ represents the auxiliary branch.

    \Remark These structures are general for all the four implemented base trackers \cite{Li2019SiamRPNEO,Wang2019Mask,Guo2021SiamGAT}.

    \myparagraph{Training Settings.} All the predictive modules need temporal video data for training. 
    However, to our disappointment, existing training pipeline \cite{Li2019SiamRPNEO} takes a detection-like paragdim. 
    Basically, the raw search patches are \textit{independently} cropped from the object center location, then the random shift, padding are applied to generated the training search patch.
    In this case, the training patches from consecutive frames actually contain no temporal information.
    
    To solve this, we construct a new pipeline termed as dynamic temporal training. The search patch from ${f_j}$-th frame is cropped around the object's center location in the previous frame $\mathcal{I}_{f_{j-1}}$, so that past motion $\mathbf{M}_{\phi(f)}$ and past search patch $\mathbf{X}_{\phi(f)}$ correspond to each other and contain real temporal information from $\mathcal{I}_{f_{j-k+1}}$ to $\mathcal{I}_{f_j}$.
    
    \Remark The new training pipeline is dynamic, \textit{i.e.}, $[{f_{j-k}}, {f_{j-k+1}}, \cdots, {f_j}]$ can be adjusted as hyper-parameters to fit different models' different latency.
    
    All the PVT++ models are optimized by AdamW \cite{Loshchilov2018AdamW}. 
    The motion predictor is trained for $100$ epochs with a base learning rate equalling to $0.03$, which is multiplied by $0.1$ at epoch $30$ and $80$. 
    The visual and multi-modal predictors are trained for $300$ epochs with a base learning rate of $0.003$, which is multiplied by $0.1$ at epoch $200$. 
    In all the four base trackers \cite{Li2019SiamRPNEO,Wang2019Mask,Guo2021SiamGAT}, $\mathcal{P}_{\mathrm{V}}$ and $\mathcal{P}_{\mathrm{MV}}$ both take the visual feature from the neck to implement vision-aided prediction.
    During joint training, the tracker backbone is fixed and the tracker neck, together with the head are freed in the first 20 epochs with a small learning rate of $10^{-5}$.
    
    A "fast" tracker may only need to predict future three frames to compensate for its latency, while a "slow" one may have to output ten future state. To make this possible, the second last layer of PVT++ predictive decoder is $N$ parallel fully connected layers for predicting $N$ future state, \textit{i.e.}, future $1\sim N$ frames. Therefore, different models vary in the pre-defined $N$ and $\Delta_f$ during training. we set $N=3, \Delta_f=[1:3]$ for SiamRPN++$_\mathrm{M}$ \cite{Li2019SiamRPNEO}, $N=12, \Delta_f=[1:12]$ for SiamRPN++$_\mathrm{R}$ \cite{Li2019SiamRPNEO}, $N=6, \Delta_f=[1:6]$ for SiamMask \cite{Wang2019Mask}, and $N=4, \Delta_f=[1:3]$ for SiamGAT \cite{Guo2021SiamGAT}. Note that these hyper-parameter are roughly determined by the averaged latency of the base trackers.

    \myparagraph{Inference Settings.} During inference, when $f_{j+1}-$th frame comes, the predictor $\mathcal{P}$ first conducts $(f_{j+1}-f_j)$ to $f_{j+1}+N$ frames prediction with $k=3$ past frames information, then the tracker processes $f_{j+1}-th$ frame and updates the history information (motion and visual).
        
    Note that we take the latency of both tracker and predictor modules into account in the online evaluation.
    
\section{Complete Notation Reference Table}\label{app:notation}
	
    We provide the important notations, their meaning, and dimension in \tref{tab:notation}, for clear reference.

\section{Visualization}\label{app:visual}

    \begin{figure*}[!t]
    \centering
    \includegraphics[width=1.7\columnwidth]{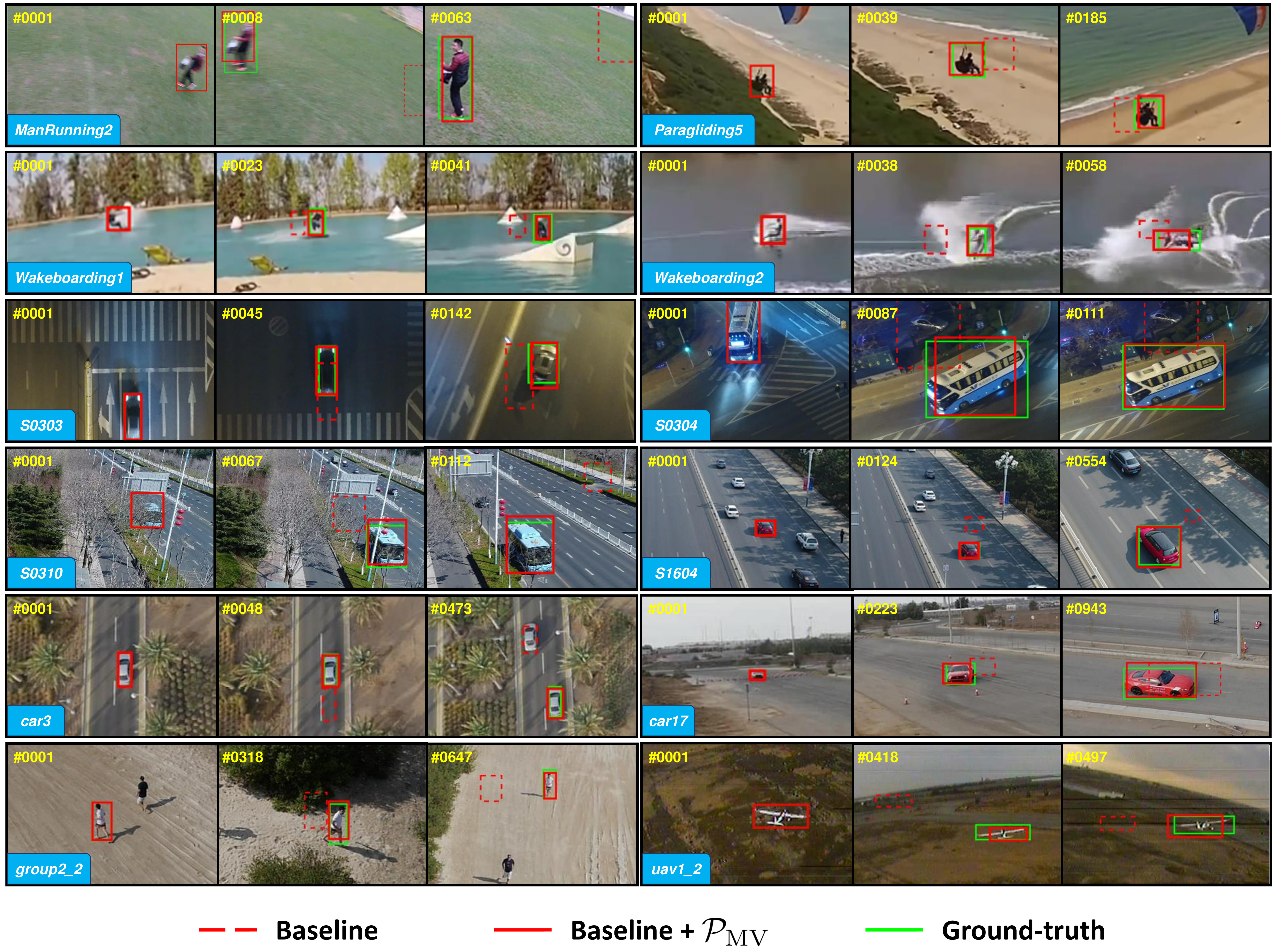}
    \vspace{-0.2cm}
    \caption{Representative sequences from authoritative UAV tracking datasets, DTB70~\cite{Li2017DTB70}, UAVDT~\cite{Du2018UAVDT}, UAV20L~\cite{Mueller2016UAV123}, and UAV123~\cite{Mueller2016UAV123}. We use dashed \textcolor{red}{red} lines to demonstrate the original trackers, which are severely affected by onboard latency. Coupled with our PVT++ ($\mathcal{P}_{\mathrm{MV}}$), the robustness can be significantly improved (solid \textcolor{red}{red} boxes). \textcolor{green}{Green} boxes denote ground-truth. Some typical sequences are also made into supplementary video for better reference.}
    \vspace{-0.3cm}
    \label{fig:visual}
    \end{figure*}

    \begin{table}[!t]
    \centering
    \setlength{\tabcolsep}{1.2mm}
    \fontsize{7}{8}\selectfont
    \caption{Efficiency and complexity comparison between PVT++ and other motion predictors \cite{gupta2018social,zhang2019srlstm,liang2019next}. Our framework is 10x$\sim$100x faster than other works.}
    \begin{threeparttable}
    \begin{tabular}{cccc|cc}
    \toprule[1.5pt]
    Input & \multicolumn{3}{c|}{Traj.} & \multicolumn{2}{c}{Traj. + RGB} \\
    \midrule
    Model & Social GAN~\cite{gupta2018social} & SR-LSTM~\cite{zhang2019srlstm} & $\mathcal{P}_\mathrm{M}$ & NEXT~\cite{liang2019next}  & $\mathcal{P}_\mathrm{MV}$ \\
    MACs  & 5.6M & 51.7M & \textbf{0.05M} & 2.7G & \textbf{1.2G} \\
    Latency (ms) & 50.2   & 652.0   & \textbf{4.2} & 181.6   & \textbf{8.6} \\
    \bottomrule[1.5pt]
    \end{tabular}
    \end{threeparttable}
    \label{tab:eff}%
    \end{table}

    \begin{table}[!t]
        \centering
        \setlength{\tabcolsep}{1.2mm}
        \fontsize{7}{8}\selectfont
        \caption{Effect of PVT++ on transformer-based trackers \cite{ye2022joint,cui2022mixformer}. Our framework can boost the perfromance by up to \textbf{40}\%.}
        \begin{threeparttable}
        \begin{tabular}{ccccc|cccc}
        \toprule[1.5pt]
        Dataset & \multicolumn{4}{c|}{DTB70}     & \multicolumn{4}{c}{UAVDT} \\
        \midrule
        Metric & \multicolumn{2}{c}{AUC@La0} & \multicolumn{2}{c|}{DP@La0} & \multicolumn{2}{c}{AUC@La0} & \multicolumn{2}{c}{DP@La0} \\
        PVT++ & \xmark  & \cmark  & \xmark  & \cmark  & \xmark  & \cmark & \xmark  & \cmark \\
        OSTrack~\cite{ye2022joint} & 0.306 & \textbf{0.400} & 0.375 & \textbf{0.535} & 0.533 & \textbf{0.626} & 0.789 & \textbf{0.839} \\
        MixFormer~\cite{cui2022mixformer} & 0.198 & \textbf{0.250} & 0.242 & \textbf{0.320} & 0.413 & \textbf{0.516} & 0.644 & \textbf{0.719} \\
        \bottomrule[1.5pt]
        \end{tabular}
        \vspace{-0.4cm}
        \end{threeparttable}
        \label{tab:tf}%
    \end{table}

    \begin{table}[!t]
        \centering
        \setlength{\tabcolsep}{1.2mm}
        \fontsize{7}{8}\selectfont
        \caption{Results comparison between two fusion strategy. $\mathcal{P}_{\mathrm{MV}}$ denotes our default PVT++, the modalities fuse after independent temporal interaction (late fusion). $\mathcal{P}^\dagger_{\mathrm{MV}}$ indicates that the two cues fuse before temporal interaction (early fusion).}
        \begin{threeparttable}
        \begin{tabular}{ccccc}
            \toprule[1.5pt]
            & \multicolumn{2}{c}{DTB70} & \multicolumn{2}{c}{UAVDT} \\
            Pred. & AUC@La0 & DP@La0 & AUC@La0 & DP@La0 \\
            \midrule
            N/A   & 0.305 & 0.387 & 0.494 & 0.719 \\
            $\mathcal{P}_{\mathrm{MV}}$ (late fuse) & \textbf{0.399} & \textbf{0.536} & \textbf{0.576} & \textbf{0.807} \\
            $\mathcal{P}^\dagger_{\mathrm{MV}}$ (early fuse) & 0.370  & 0.498 & 0.571 & 0.800 \\
            \bottomrule[1.5pt]
        \end{tabular}%
        \end{threeparttable}
        \label{tab:fuse}%
    \end{table}%
    We present some typical tracking visualization in \fref{fig:visual}. The sequences, \textit{ManRunning2}, \textit{Paragliding5}, \textit{Wakeboarding1}, and \textit{Wakeboarding2} are from DTB70~\cite{Li2017DTB70}.\textit{S0303}, \textit{S0304}, \textit{S0310}, and \textit{S1604} are from UAVDT~\cite{Du2018UAVDT}. In UAV20L and UAV123~\cite{Mueller2016UAV123}, we also present \textit{car3}, \textit{car17}, \textit{group2\_2}, and \textit{uav1\_2}. With extremely limited onboard computation, the original trackers (red dashed boxes) will easily fail due to high latency. Once coupled with our PVT++ ($\mathcal{P}_{\mathrm{MV}}$), the models (solid red boxes) are much more robust. We use greed boxes to denote ground-truth.
    
\section{Prediction Quantitative Comparison}\label{app:allkf}

 \begin{table*}[!t]
    \centering
    \setlength{\tabcolsep}{1.4mm}
    \fontsize{8}{9}\selectfont
    \caption{Attribute-based analysis of the three trackers with PVT++ models in DTB70~\cite{Li2017DTB70} dataset.}
    \begin{threeparttable}
        \begin{tabular}{cc|cccc|cccc|cccc}
            \toprule[1.5pt]
            \multicolumn{2}{c|}{Tracker} & \multicolumn{4}{c|}{\makecell[c]{SiamRPN++$_{\mathrm{M}}$\\(21FPS)}} & \multicolumn{4}{c|}{\makecell[c]{SiamRPN++$_{\mathrm{R}}$\\(5FPS)}} & \multicolumn{4}{c}{\makecell[c]{SiamMask\\(12FPS)}} \\
            \midrule
            Metric & Att.  & N/A  & $\mathcal{P}_\mathrm{M}$ & $\mathcal{P}_\mathrm{V}$ & $\mathcal{P}_\mathrm{MV}$ & N/A  & $\mathcal{P}_\mathrm{M}$ & $\mathcal{P}_\mathrm{V}$ & $\mathcal{P}_\mathrm{MV}$ & N/A  & $\mathcal{P}_\mathrm{M}$ & $\mathcal{P}_\mathrm{V}$ & \multicolumn{1}{c}{$\mathcal{P}_\mathrm{MV}$} \\
            \multirow{11}[1]{*}{AUC@La0} & ARV   & 0.330 & 0.386 & 0.349 & \cellcolor[rgb]{ .806,  .802,  .802}0.418 & 0.156 & 0.233 & 0.214 & \cellcolor[rgb]{ .806,  .802,  .802}0.253 & 0.247 & 0.375 & 0.291 & \cellcolor[rgb]{ .806,  .802,  .802}0.393 \\
            & BC    & 0.257 & \cellcolor[rgb]{ .806,  .802,  .802}0.330 & 0.276 & 0.319 & 0.079 & 0.077 & 0.102 & \cellcolor[rgb]{ .806,  .802,  .802}0.102 & 0.168 & \cellcolor[rgb]{ .806,  .802,  .802}0.264 & 0.202 & 0.167 \\
            & DEF   & 0.357 & 0.410 & 0.358 & \cellcolor[rgb]{ .806,  .802,  .802}0.438 & 0.144 & 0.217 & 0.198 & \cellcolor[rgb]{ .806,  .802,  .802}0.241 & 0.253 & \cellcolor[rgb]{ .806,  .802,  .802}0.398 & 0.287 & 0.364 \\
            & FCM   & 0.277 & 0.373 & 0.333 & \cellcolor[rgb]{ .806,  .802,  .802}0.376 & 0.091 & \cellcolor[rgb]{ .806,  .802,  .802}0.144 & 0.122 & 0.138 & 0.195 & \cellcolor[rgb]{ .806,  .802,  .802}0.327 & 0.258 & 0.301 \\
            & IPR   & 0.302 & 0.368 & 0.324 & \cellcolor[rgb]{ .806,  .802,  .802}0.387 & 0.133 & 0.187 & 0.169 & \cellcolor[rgb]{ .806,  .802,  .802}0.204 & 0.217 & \cellcolor[rgb]{ .806,  .802,  .802}0.346 & 0.256 & 0.316 \\
            & MB    & 0.198 & 0.305 & 0.277 & \cellcolor[rgb]{ .806,  .802,  .802}0.321 & 0.056 & 0.073 & 0.069 & \cellcolor[rgb]{ .806,  .802,  .802}0.085 & 0.147 & 0.236 & 0.187 & \cellcolor[rgb]{ .806,  .802,  .802}0.254 \\
            & OCC   & 0.280 & \cellcolor[rgb]{ .806,  .802,  .802}0.337 & 0.281 & 0.304 & 0.149 & 0.214 & 0.204 & \cellcolor[rgb]{ .806,  .802,  .802}0.224 & 0.233 & \cellcolor[rgb]{ .806,  .802,  .802}0.290 & 0.285 & 0.274 \\
            & OPR   & 0.278 & 0.314 & 0.334 & \cellcolor[rgb]{ .806,  .802,  .802}0.439 & 0.161 & 0.158 & 0.208 & \cellcolor[rgb]{ .806,  .802,  .802}0.225 & 0.202 & 0.360 & 0.265 & \cellcolor[rgb]{ .806,  .802,  .802}0.362 \\
            & OV    & 0.292 & \cellcolor[rgb]{ .806,  .802,  .802}0.405 & 0.372 & 0.399 & 0.054 & 0.099 & 0.076 & \cellcolor[rgb]{ .806,  .802,  .802}0.102 & 0.168 & 0.227 & 0.258 & \cellcolor[rgb]{ .806,  .802,  .802}0.289 \\
            & SV    & 0.354 & 0.470 & 0.419 & \cellcolor[rgb]{ .806,  .802,  .802}0.489 & 0.145 & 0.187 & 0.192 & \cellcolor[rgb]{ .806,  .802,  .802}0.220 & 0.278 & \cellcolor[rgb]{ .806,  .802,  .802}0.435 & 0.347 & 0.418 \\
            & SOA   & 0.238 & 0.301 & 0.261 & \cellcolor[rgb]{ .806,  .802,  .802}0.302 & 0.140 & 0.196 & 0.184 & \cellcolor[rgb]{ .806,  .802,  .802}0.200 & 0.227 & \cellcolor[rgb]{ .806,  .802,  .802}0.326 & 0.275 & 0.315 \\
            \midrule
            \multirow{11}[2]{*}{DP@La0} & ARV   & 0.340 & 0.466 & 0.385 & \cellcolor[rgb]{ .806,  .802,  .802}0.498 & 0.101 & 0.220 & 0.171 & \cellcolor[rgb]{ .806,  .802,  .802}0.234 & 0.247 & \cellcolor[rgb]{ .806,  .802,  .802}0.474 & 0.333 & 0.472 \\
            & BC    & 0.352 & 0.477 & 0.396 & \cellcolor[rgb]{ .806,  .802,  .802}0.498 & 0.118 & 0.106 & \cellcolor[rgb]{ .806,  .802,  .802}0.141 & 0.139 & 0.228 & \cellcolor[rgb]{ .806,  .802,  .802}0.385 & 0.291 & 0.237 \\
            & DEF   & 0.374 & 0.512 & 0.398 & \cellcolor[rgb]{ .806,  .802,  .802}0.525 & 0.083 & 0.203 & 0.144 & \cellcolor[rgb]{ .806,  .802,  .802}0.214 & 0.246 & \cellcolor[rgb]{ .806,  .802,  .802}0.509 & 0.326 & 0.449 \\
            & FCM   & 0.363 & 0.517 & 0.470 & \cellcolor[rgb]{ .806,  .802,  .802}0.525 & 0.106 & 0.188 & 0.156 & \cellcolor[rgb]{ .806,  .802,  .802}0.171 & 0.241 & \cellcolor[rgb]{ .806,  .802,  .802}0.456 & 0.353 & 0.414 \\
            & IPR   & 0.349 & 0.475 & 0.398 & \cellcolor[rgb]{ .806,  .802,  .802}0.495 & 0.124 & 0.212 & 0.170 & \cellcolor[rgb]{ .806,  .802,  .802}0.224 & 0.236 & \cellcolor[rgb]{ .806,  .802,  .802}0.454 & 0.310 & 0.400 \\
            & MB    & 0.246 & 0.418 & 0.379 & \cellcolor[rgb]{ .806,  .802,  .802}0.453 & 0.051 & \cellcolor[rgb]{ .806,  .802,  .802}0.110 & 0.090 & 0.088 & 0.167 & \cellcolor[rgb]{ .806,  .802,  .802}0.349 & 0.248 & 0.327 \\
            & OCC   & 0.408 & \cellcolor[rgb]{ .806,  .802,  .802}0.496 & 0.426 & 0.459 & 0.223 & 0.327 & 0.316 & \cellcolor[rgb]{ .806,  .802,  .802}0.344 & 0.361 & \cellcolor[rgb]{ .806,  .802,  .802}0.439 & 0.458 & 0.404 \\
            & OPR   & 0.213 & 0.312 & 0.317 & \cellcolor[rgb]{ .806,  .802,  .802}0.453 & 0.083 & 0.083 & 0.113 & \cellcolor[rgb]{ .806,  .802,  .802}0.127 & 0.128 & \cellcolor[rgb]{ .806,  .802,  .802}0.382 & 0.224 & 0.357 \\
            & OV    & 0.413 & \cellcolor[rgb]{ .806,  .802,  .802}0.590 & 0.564 & 0.586 & 0.062 & \cellcolor[rgb]{ .806,  .802,  .802}0.166 & 0.101 & 0.161 & 0.222 & 0.363 & 0.385 & \cellcolor[rgb]{ .806,  .802,  .802}0.439 \\
            & SV    & 0.366 & 0.569 & 0.467 & \cellcolor[rgb]{ .806,  .802,  .802}0.569 & 0.123 & 0.186 & 0.180 & \cellcolor[rgb]{ .806,  .802,  .802}0.208 & 0.287 & \cellcolor[rgb]{ .806,  .802,  .802}0.528 & 0.402 & 0.492 \\
            & SOA   & 0.333 & 0.432 & 0.379 & \cellcolor[rgb]{ .806,  .802,  .802}0.447 & 0.217 & \cellcolor[rgb]{ .806,  .802,  .802}0.306 & 0.295 & 0.302 & 0.340 & \cellcolor[rgb]{ .806,  .802,  .802}0.479 & 0.429 & 0.462 \\
            \bottomrule[1.5pt]
        \end{tabular}%
    \end{threeparttable}
    \label{tab:att_DTB}%
\end{table*}%

\begin{table*}[!t]
    \centering
    \setlength{\tabcolsep}{1.2mm}
    \fontsize{8}{9}\selectfont
    \caption{Attribute-based analysis of the three trackers with PVT++ models in UAVDT~\cite{Du2018UAVDT} dataset.}
    \begin{threeparttable}
    \begin{tabular}{cc|cccc|cccc|cccc}
    \toprule[1.5pt]
    \multicolumn{2}{c|}{Tracker} & \multicolumn{4}{c|}{\makecell[c]{SiamRPN++$_{\mathrm{M}}$\\(21FPS)}} & \multicolumn{4}{c|}{\makecell[c]{SiamRPN++$_{\mathrm{R}}$\\(5FPS)}} & \multicolumn{4}{c}{\makecell[c]{SiamMask\\(12FPS)}} \\
    \midrule
    Metric & Att.  & N/A   & $\mathcal{P}_\mathrm{M}$ & $\mathcal{P}_\mathrm{V}$ & $\mathcal{P}_\mathrm{MV}$ & N/A   & $\mathcal{P}_\mathrm{M}$ & $\mathcal{P}_\mathrm{V}$ & $\mathcal{P}_\mathrm{MV}$ & N/A   & $\mathcal{P}_\mathrm{M}$ & $\mathcal{P}_\mathrm{V}$ & \multicolumn{1}{c}{$\mathcal{P}_\mathrm{MV}$} \\
    \multirow{8}[1]{*}{AUC@La0} & BC    & 0.448 & 0.461 & 0.504 & \cellcolor[rgb]{ .806,  .802,  .802}0.505 & 0.332 & 0.410 & 0.375 & \cellcolor[rgb]{ .806,  .802,  .802}0.445 & 0.404 & 0.465 & 0.488 & \cellcolor[rgb]{ .806,  .802,  .802}0.520 \\
          & CR    & 0.450 & 0.495 & 0.520 & \cellcolor[rgb]{ .806,  .802,  .802}0.535 & 0.296 & 0.371 & 0.402 & \cellcolor[rgb]{ .806,  .802,  .802}0.452 & 0.425 & 0.503 & 0.498 & \cellcolor[rgb]{ .806,  .802,  .802}0.522 \\
          & OR    & 0.438 & 0.481 & 0.538 & \cellcolor[rgb]{ .806,  .802,  .802}0.549 & 0.318 & 0.389 & 0.416 & \cellcolor[rgb]{ .806,  .802,  .802}0.477 & 0.404 & 0.491 & 0.504 & \cellcolor[rgb]{ .806,  .802,  .802}0.541 \\
          & SO    & 0.494 & \cellcolor[rgb]{ .806,  .802,  .802}0.549 & 0.525 & 0.545 & 0.318 & 0.420 & 0.361 & \cellcolor[rgb]{ .806,  .802,  .802}0.457 & 0.468 & 0.536 & 0.495 & \cellcolor[rgb]{ .806,  .802,  .802}0.540 \\
          & IV    & 0.539 & 0.578 & 0.588 & \cellcolor[rgb]{ .806,  .802,  .802}0.599 & 0.382 & 0.495 & 0.459 & \cellcolor[rgb]{ .806,  .802,  .802}0.537 & 0.475 & 0.558 & 0.563 & \cellcolor[rgb]{ .806,  .802,  .802}0.596 \\
          & OB    & 0.525 & 0.542 & 0.568 & \cellcolor[rgb]{ .806,  .802,  .802}0.589 & 0.382 & 0.460 & 0.408 & \cellcolor[rgb]{ .806,  .802,  .802}0.498 & 0.471 & 0.542 & 0.527 & \cellcolor[rgb]{ .806,  .802,  .802}0.560 \\
          & SV    & 0.490 & 0.505 & 0.584 & \cellcolor[rgb]{ .806,  .802,  .802}0.586 & 0.366 & 0.422 & 0.406 & \cellcolor[rgb]{ .806,  .802,  .802}0.484 & 0.438 & 0.526 & 0.541 & \cellcolor[rgb]{ .806,  .802,  .802}0.566 \\
          & LO    & 0.422 & \cellcolor[rgb]{ .806,  .802,  .802}0.521 & 0.436 & 0.511 & 0.320 & 0.379 & 0.368 & \cellcolor[rgb]{ .806,  .802,  .802}0.429 & 0.389 & 0.421 & 0.494 & \cellcolor[rgb]{ .806,  .802,  .802}0.520 \\
    \midrule
    \multirow{8}[2]{*}{DP@La0} & BC    & 0.659 & 0.666 & \cellcolor[rgb]{ .806,  .802,  .802}0.733 & 0.727 & 0.591 & 0.637 & 0.647 & \cellcolor[rgb]{ .806,  .802,  .802}0.671 & 0.628 & 0.672 & 0.718 & \cellcolor[rgb]{ .806,  .802,  .802}0.731 \\
          & CR    & 0.643 & 0.684 & 0.720 & \cellcolor[rgb]{ .806,  .802,  .802}0.732 & 0.462 & 0.585 & 0.572 & \cellcolor[rgb]{ .806,  .802,  .802}0.645 & 0.620 & 0.702 & 0.696 & \cellcolor[rgb]{ .806,  .802,  .802}0.712 \\
          & OR    & 0.638 & 0.681 & 0.753 & \cellcolor[rgb]{ .806,  .802,  .802}0.764 & 0.515 & 0.619 & 0.606 & \cellcolor[rgb]{ .806,  .802,  .802}0.688 & 0.612 & 0.709 & 0.723 & \cellcolor[rgb]{ .806,  .802,  .802}0.752 \\
          & SO    & 0.779 & \cellcolor[rgb]{ .806,  .802,  .802}0.815 & 0.793 & 0.814 & 0.645 & 0.711 & 0.706 & \cellcolor[rgb]{ .806,  .802,  .802}0.759 & 0.803 & 0.818 & 0.787 & \cellcolor[rgb]{ .806,  .802,  .802}0.819 \\
          & IV    & 0.777 & 0.811 & 0.835 & \cellcolor[rgb]{ .806,  .802,  .802}0.848 & 0.657 & 0.747 & 0.755 & \cellcolor[rgb]{ .806,  .802,  .802}0.801 & 0.743 & 0.797 & 0.817 & \cellcolor[rgb]{ .806,  .802,  .802}0.829 \\
          & OB    & 0.772 & 0.778 & 0.822 & \cellcolor[rgb]{ .806,  .802,  .802}0.846 & 0.676 & 0.714 & 0.700 & \cellcolor[rgb]{ .806,  .802,  .802}0.766 & 0.756 & 0.802 & 0.801 & \cellcolor[rgb]{ .806,  .802,  .802}0.813 \\
          & SV    & 0.680 & 0.691 & \cellcolor[rgb]{ .806,  .802,  .802}0.796 & 0.794 & 0.581 & 0.618 & 0.622 & \cellcolor[rgb]{ .806,  .802,  .802}0.684 & 0.650 & 0.729 & 0.763 & \cellcolor[rgb]{ .806,  .802,  .802}0.783 \\
          & LO    & 0.569 & \cellcolor[rgb]{ .806,  .802,  .802}0.717 & 0.585 & 0.694 & 0.504 & 0.554 & 0.566 & \cellcolor[rgb]{ .806,  .802,  .802}0.608 & 0.571 & 0.590 & 0.696 & \cellcolor[rgb]{ .806,  .802,  .802}0.711 \\
    \bottomrule[1.5pt]
    \end{tabular}%
    \end{threeparttable}
         \label{tab:att_DT}%
\end{table*}%
	To provide a thorough quantitative comparison of the predictor performance, we reported the results per dataset in \tref{tab:full_kf}. We observe that for different tracker models in various benchmarks, PVT++ is more robust than prior solutions~\cite{Li2020PredictiveVT,Li2020TowardsSP}. Compared with learnable KFs, KF$^\dagger$ and KF$\ddagger$, our PVT++ holds obvious advantage by virtue of the visual cue and joint learning. 

\section{Effect on Transformer-based Trackers}\label{app:tf}
	
    For transformer-based trackers, MixFormer~\cite{cui2022mixformer} and OSTrack~\cite{ye2022joint} ($\sim$6 and $\sim$10 FPS onboard), PVT++ yields up to $40$\% improvement as shown in \tref{tab:tf}.

\section{Efficiency and Complexity Comparison}\label{app:eff}
	
    PVT++ is a lightweight plug-and-play framework designed for latency-aware tracking, while most existing trajectory predictors are computationally heavy. As in \tref{tab:eff}, PVT++ is 10x$\sim$100x faster than existing trajectory predictors and introduces much less extra latency onboard.

\section{Fusion Strategy Comparison}\label{app:earlyfus}
    
    As introduced in the paper, inside PVT++, the three modules, \textit{Feature encoder}, \textit{temporal interaction}, and \textit{predictive decoder} run one after another. For the default setting, the fusion of the motion and visual cues happens after \textit{temporal interaction}, using the concatenate function. 
    Here, we also tried to integrate the two modality earlier before \textit{temporal interaction} and right after \textit{feature encoder}, still adopting concatenation. The results comparison of two strategies is shown in \tref{tab:fuse}, where we find both are effective and the late fusion is better.

\section{Full Attribute-based Analysis}\label{app:att}

    \begin{table*}[!t]
    	\centering
    	\setlength{\tabcolsep}{1.2mm}
    	\fontsize{8}{9}\selectfont
    	\caption{Attribute-based analysis of the three trackers with PVT++ models in UAV20L~\cite{Mueller2016UAV123} dataset.}
    	\begin{threeparttable}
        \begin{tabular}{cc|cccc|cccc|cccc}
        \toprule[1.5pt]
        \multicolumn{2}{c|}{Tracker} & \multicolumn{4}{c|}{\makecell[c]{SiamRPN++$_{\mathrm{M}}$\\(21FPS)}} & \multicolumn{4}{c|}{\makecell[c]{SiamRPN++$_{\mathrm{R}}$\\(5FPS)}} & \multicolumn{4}{c}{\makecell[c]{SiamMask\\(12FPS)}} \\
        \midrule
        Metric & Att.  & N/A   & $\mathcal{P}_\mathrm{M}$ & $\mathcal{P}_\mathrm{V}$ & $\mathcal{P}_\mathrm{MV}$ & N/A   & $\mathcal{P}_\mathrm{M}$ & $\mathcal{P}_\mathrm{V}$ & $\mathcal{P}_\mathrm{MV}$ & N/A   & $\mathcal{P}_\mathrm{M}$ & $\mathcal{P}_\mathrm{V}$ & $\mathcal{P}_\mathrm{MV}$ \\
        \multirow{12}[1]{*}{AUC@La0} & SV    & 0.437 & 0.470 & 0.483 & \cellcolor[rgb]{ .806,  .802,  .802}0.500 & 0.300 & 0.395 & 0.392 & \cellcolor[rgb]{ .806,  .802,  .802}0.410 & 0.395 & 0.437 & 0.420 & \cellcolor[rgb]{ .806,  .802,  .802}0.461 \\
              & ARC   & 0.425 & 0.411 & 0.438 & \cellcolor[rgb]{ .806,  .802,  .802}0.451 & 0.291 & 0.352 & 0.360 & \cellcolor[rgb]{ .806,  .802,  .802}0.371 & 0.373 & 0.409 & 0.392 & \cellcolor[rgb]{ .806,  .802,  .802}0.438 \\
              & LR    & 0.267 & \cellcolor[rgb]{ .806,  .802,  .802}0.354 & 0.344 & 0.352 & 0.215 & \cellcolor[rgb]{ .806,  .802,  .802}0.295 & 0.276 & 0.279 & 0.244 & 0.263 & 0.275 & \cellcolor[rgb]{ .806,  .802,  .802}0.290 \\
              & FM    & 0.410 & 0.357 & 0.394 & \cellcolor[rgb]{ .806,  .802,  .802}0.418 & 0.269 & 0.304 & \cellcolor[rgb]{ .806,  .802,  .802}0.325 & 0.315 & 0.319 & 0.375 & 0.329 & \cellcolor[rgb]{ .806,  .802,  .802}0.442 \\
              & FOC   & 0.256 & \cellcolor[rgb]{ .806,  .802,  .802}0.272 & 0.234 & 0.241 & 0.170 & \cellcolor[rgb]{ .806,  .802,  .802}0.227 & 0.184 & 0.164 & 0.221 & 0.237 & 0.231 & \cellcolor[rgb]{ .806,  .802,  .802}0.255 \\
              & POC   & 0.418 & \cellcolor[rgb]{ .806,  .802,  .802}0.480 & 0.463 & 0.478 & 0.286 & 0.379 & 0.380 & \cellcolor[rgb]{ .806,  .802,  .802}0.396 & 0.378 & 0.417 & 0.430 & \cellcolor[rgb]{ .806,  .802,  .802}0.441 \\
              & OV    & 0.438 & \cellcolor[rgb]{ .806,  .802,  .802}0.512 & 0.476 & 0.492 & 0.272 & 0.356 & 0.394 & \cellcolor[rgb]{ .806,  .802,  .802}0.405 & 0.377 & 0.428 & 0.448 & \cellcolor[rgb]{ .806,  .802,  .802}0.462 \\
              & BC    & 0.225 & \cellcolor[rgb]{ .806,  .802,  .802}0.258 & 0.229 & 0.250 & 0.119 & \cellcolor[rgb]{ .806,  .802,  .802}0.215 & 0.153 & 0.159 & 0.189 & 0.198 & 0.210 & \cellcolor[rgb]{ .806,  .802,  .802}0.210 \\
              & IV    & 0.452 & 0.414 & 0.470 & \cellcolor[rgb]{ .806,  .802,  .802}0.491 & 0.303 & 0.393 & 0.379 & \cellcolor[rgb]{ .806,  .802,  .802}0.403 & 0.426 & 0.437 & 0.382 & \cellcolor[rgb]{ .806,  .802,  .802}0.443 \\
              & VC    & 0.472 & 0.450 & 0.466 & \cellcolor[rgb]{ .806,  .802,  .802}0.488 & 0.302 & 0.339 & 0.377 & \cellcolor[rgb]{ .806,  .802,  .802}0.384 & 0.395 & 0.436 & 0.420 & \cellcolor[rgb]{ .806,  .802,  .802}0.475 \\
              & CM    & 0.431 & 0.463 & 0.475 & \cellcolor[rgb]{ .806,  .802,  .802}0.491 & 0.297 & \cellcolor[rgb]{ .806,  .802,  .802}0.393 & 0.388 & 0.406 & 0.391 & 0.432 & 0.412 & \cellcolor[rgb]{ .806,  .802,  .802}0.452 \\
              & SO    & 0.482 & 0.519 & 0.557 & \cellcolor[rgb]{ .806,  .802,  .802}0.567 & 0.399 & \cellcolor[rgb]{ .806,  .802,  .802}0.531 & 0.477 & 0.491 & 0.487 & \cellcolor[rgb]{ .806,  .802,  .802}0.519 & 0.438 & 0.492 \\
        \midrule
        \multirow{12}[2]{*}{DP@La0} & SV    & 0.600 & 0.630 & 0.662 & \cellcolor[rgb]{ .806,  .802,  .802}0.683 & 0.417 & 0.544 & 0.536 & \cellcolor[rgb]{ .806,  .802,  .802}0.556 & 0.552 & 0.588 & 0.581 & \cellcolor[rgb]{ .806,  .802,  .802}0.627 \\
              & ARC   & 0.591 & 0.562 & 0.606 & \cellcolor[rgb]{ .806,  .802,  .802}0.624 & 0.408 & 0.487 & 0.486 & \cellcolor[rgb]{ .806,  .802,  .802}0.503 & 0.524 & 0.558 & 0.550 & \cellcolor[rgb]{ .806,  .802,  .802}0.603 \\
              & LR    & 0.444 & 0.545 & 0.539 & \cellcolor[rgb]{ .806,  .802,  .802}0.548 & 0.388 & \cellcolor[rgb]{ .806,  .802,  .802}0.483 & 0.465 & 0.456 & 0.422 & 0.414 & 0.458 & \cellcolor[rgb]{ .806,  .802,  .802}0.465 \\
              & FM    & 0.631 & 0.548 & 0.595 & \cellcolor[rgb]{ .806,  .802,  .802}0.625 & 0.417 & 0.464 & \cellcolor[rgb]{ .806,  .802,  .802}0.495 & 0.476 & 0.518 & 0.573 & 0.524 & \cellcolor[rgb]{ .806,  .802,  .802}0.667 \\
              & FOC   & 0.469 & \cellcolor[rgb]{ .806,  .802,  .802}0.473 & 0.436 & 0.428 & 0.358 & \cellcolor[rgb]{ .806,  .802,  .802}0.423 & 0.358 & 0.324 & 0.425 & 0.420 & 0.431 & \cellcolor[rgb]{ .806,  .802,  .802}0.459 \\
              & POC   & 0.585 & 0.654 & 0.648 & \cellcolor[rgb]{ .806,  .802,  .802}0.669 & 0.410 & 0.530 & 0.531 & \cellcolor[rgb]{ .806,  .802,  .802}0.548 & 0.540 & 0.570 & 0.606 & \cellcolor[rgb]{ .806,  .802,  .802}0.613 \\
              & OV    & 0.597 & \cellcolor[rgb]{ .806,  .802,  .802}0.683 & 0.658 & 0.679 & 0.356 & 0.473 & 0.518 & \cellcolor[rgb]{ .806,  .802,  .802}0.540 & 0.529 & 0.578 & 0.618 & \cellcolor[rgb]{ .806,  .802,  .802}0.630 \\
              & BC    & 0.426 & \cellcolor[rgb]{ .806,  .802,  .802}0.440 & 0.399 & 0.434 & 0.284 & \cellcolor[rgb]{ .806,  .802,  .802}0.398 & 0.304 & 0.295 & 0.378 & 0.349 & \cellcolor[rgb]{ .806,  .802,  .802}0.390 & 0.385 \\
              & IV    & 0.628 & 0.560 & 0.649 & \cellcolor[rgb]{ .806,  .802,  .802}0.686 & 0.428 & \cellcolor[rgb]{ .806,  .802,  .802}0.551 & 0.503 & 0.539 & 0.595 & 0.590 & 0.545 & \cellcolor[rgb]{ .806,  .802,  .802}0.617 \\
              & VC    & 0.616 & 0.571 & 0.605 & \cellcolor[rgb]{ .806,  .802,  .802}0.631 & 0.364 & 0.420 & 0.452 & \cellcolor[rgb]{ .806,  .802,  .802}0.477 & 0.518 & 0.551 & 0.546 & \cellcolor[rgb]{ .806,  .802,  .802}0.611 \\
              & CM    & 0.599 & 0.629 & 0.660 & \cellcolor[rgb]{ .806,  .802,  .802}0.681 & 0.417 & 0.544 & 0.534 & \cellcolor[rgb]{ .806,  .802,  .802}0.553 & 0.550 & 0.588 & 0.580 & \cellcolor[rgb]{ .806,  .802,  .802}0.626 \\
              & SO    & 0.604 & 0.652 & 0.719 & \cellcolor[rgb]{ .806,  .802,  .802}0.734 & 0.498 & \cellcolor[rgb]{ .806,  .802,  .802}0.645 & 0.594 & 0.609 & 0.610 & \cellcolor[rgb]{ .806,  .802,  .802}0.648 & 0.559 & 0.619 \\
        \bottomrule[1.5pt]
        \end{tabular}%
    	\end{threeparttable}
     \label{tab:att_20L}%
    \end{table*}%
    
    \begin{table*}[!t]
    	\centering
    	\setlength{\tabcolsep}{1.2mm}
    	\fontsize{8}{9}\selectfont
    	\caption{Attribute-based analysis of the three trackers with PVT++ models in UAV123~\cite{Mueller2016UAV123} dataset.}
    	\begin{threeparttable}
        \begin{tabular}{cc|cccc|cccc|cccc}
        \toprule[1.5pt]
        \multicolumn{2}{c|}{Tracker} & \multicolumn{4}{c|}{\makecell[c]{SiamRPN++$_{\mathrm{M}}$\\(21FPS)}} & \multicolumn{4}{c|}{\makecell[c]{SiamRPN++$_{\mathrm{R}}$\\(5FPS)}} & \multicolumn{4}{c}{\makecell[c]{SiamMask\\(12FPS)}} \\
        \midrule
        Metric & Att.  & N/A   & $\mathcal{P}_\mathrm{M}$ & $\mathcal{P}_\mathrm{V}$ & $\mathcal{P}_\mathrm{MV}$ & N/A   & $\mathcal{P}_\mathrm{M}$ & $\mathcal{P}_\mathrm{V}$ & $\mathcal{P}_\mathrm{MV}$ & N/A   & $\mathcal{P}_\mathrm{M}$ & $\mathcal{P}_\mathrm{V}$ & $\mathcal{P}_\mathrm{MV}$ \\
        \multirow{12}[1]{*}{AUC@La0} & SV    & 0.456 & \cellcolor[rgb]{ .806,  .802,  .802}0.518 & 0.488 & 0.514 & 0.338 & 0.423 & 0.383 & \cellcolor[rgb]{ .806,  .802,  .802}0.427 & 0.420 & 0.509 & 0.480 & \cellcolor[rgb]{ .806,  .802,  .802}0.518 \\
              & ARC   & 0.413 & \cellcolor[rgb]{ .806,  .802,  .802}0.496 & 0.468 & 0.491 & 0.315 & 0.402 & 0.365 & \cellcolor[rgb]{ .806,  .802,  .802}0.406 & 0.398 & 0.498 & 0.467 & \cellcolor[rgb]{ .806,  .802,  .802}0.510 \\
              & LR    & 0.291 & \cellcolor[rgb]{ .806,  .802,  .802}0.357 & 0.328 & 0.350 & 0.179 & \cellcolor[rgb]{ .806,  .802,  .802}0.264 & 0.214 & 0.256 & 0.257 & \cellcolor[rgb]{ .806,  .802,  .802}0.364 & 0.324 & 0.257 \\
              & FM    & 0.373 & 0.430 & 0.461 & \cellcolor[rgb]{ .806,  .802,  .802}0.482 & 0.261 & 0.316 & 0.307 & \cellcolor[rgb]{ .806,  .802,  .802}0.341 & 0.333 & 0.425 & 0.422 & \cellcolor[rgb]{ .806,  .802,  .802}0.447 \\
              & FOC   & 0.254 & \cellcolor[rgb]{ .806,  .802,  .802}0.317 & 0.270 & 0.306 & 0.191 & \cellcolor[rgb]{ .806,  .802,  .802}0.251 & 0.214 & 0.246 & 0.242 & \cellcolor[rgb]{ .806,  .802,  .802}0.325 & 0.284 & 0.303 \\
              & POC   & 0.401 & 0.436 & 0.402 & \cellcolor[rgb]{ .806,  .802,  .802}0.446 & 0.284 & 0.373 & 0.335 & \cellcolor[rgb]{ .806,  .802,  .802}0.374 & 0.363 & 0.449 & 0.426 & \cellcolor[rgb]{ .806,  .802,  .802}0.459 \\
              & OV    & 0.442 & 0.489 & 0.488 & \cellcolor[rgb]{ .806,  .802,  .802}0.516 & 0.289 & 0.394 & 0.368 & \cellcolor[rgb]{ .806,  .802,  .802}0.407 & 0.403 & \cellcolor[rgb]{ .806,  .802,  .802}0.504 & 0.476 & 0.492 \\
              & BC    & 0.254 & 0.293 & 0.247 & \cellcolor[rgb]{ .806,  .802,  .802}0.296 & 0.188 & \cellcolor[rgb]{ .806,  .802,  .802}0.258 & 0.215 & 0.247 & 0.248 & \cellcolor[rgb]{ .806,  .802,  .802}0.360 & 0.307 & 0.309 \\
              & IV    & 0.365 & 0.421 & 0.423 & \cellcolor[rgb]{ .806,  .802,  .802}0.465 & 0.310 & 0.379 & 0.352 & \cellcolor[rgb]{ .806,  .802,  .802}0.381 & 0.378 & \cellcolor[rgb]{ .806,  .802,  .802}0.480 & 0.441 & 0.466 \\
              & VC    & 0.459 & 0.552 & 0.506 & \cellcolor[rgb]{ .806,  .802,  .802}0.558 & 0.322 & 0.409 & 0.387 & \cellcolor[rgb]{ .806,  .802,  .802}0.432 & 0.407 & 0.534 & 0.499 & \cellcolor[rgb]{ .806,  .802,  .802}0.548 \\
              & CM    & 0.466 & \cellcolor[rgb]{ .806,  .802,  .802}0.542 & 0.514 & 0.535 & 0.319 & 0.421 & 0.381 & \cellcolor[rgb]{ .806,  .802,  .802}0.422 & 0.420 & \cellcolor[rgb]{ .806,  .802,  .802}0.529 & 0.502 & 0.522 \\
              & SO    & 0.478 & \cellcolor[rgb]{ .806,  .802,  .802}0.497 & 0.444 & 0.459 & 0.362 & \cellcolor[rgb]{ .806,  .802,  .802}0.462 & 0.382 & 0.435 & 0.434 & 0.492 & 0.464 & \cellcolor[rgb]{ .806,  .802,  .802}0.514 \\
        \midrule
        \multirow{12}[2]{*}{DP@La0} & SV    & 0.657 & \cellcolor[rgb]{ .806,  .802,  .802}0.714 & 0.679 & 0.710 & 0.488 & \cellcolor[rgb]{ .806,  .802,  .802}0.599 & 0.594 & 0.537 & 0.614 & 0.711 & 0.671 & \cellcolor[rgb]{ .806,  .802,  .802}0.720 \\
              & ARC   & 0.602 & \cellcolor[rgb]{ .806,  .802,  .802}0.689 & 0.651 & 0.678 & 0.453 & \cellcolor[rgb]{ .806,  .802,  .802}0.575 & 0.502 & 0.561 & 0.588 & 0.701 & 0.656 & \cellcolor[rgb]{ .806,  .802,  .802}0.715 \\
              & LR    & 0.548 & \cellcolor[rgb]{ .806,  .802,  .802}0.595 & 0.568 & 0.586 & 0.392 & \cellcolor[rgb]{ .806,  .802,  .802}0.488 & 0.471 & 0.438 & 0.510 & 0.621 & 0.554 & \cellcolor[rgb]{ .806,  .802,  .802}0.637 \\
              & FM    & 0.517 & 0.591 & 0.617 & \cellcolor[rgb]{ .806,  .802,  .802}0.646 & 0.323 & 0.417 & 0.368 & \cellcolor[rgb]{ .806,  .802,  .802}0.429 & 0.450 & 0.588 & 0.564 & \cellcolor[rgb]{ .806,  .802,  .802}0.609 \\
              & FOC   & 0.497 & \cellcolor[rgb]{ .806,  .802,  .802}0.550 & 0.489 & 0.533 & 0.387 & \cellcolor[rgb]{ .806,  .802,  .802}0.460 & 0.406 & 0.448 & 0.460 & \cellcolor[rgb]{ .806,  .802,  .802}0.569 & 0.505 & 0.541 \\
              & POC   & 0.614 & 0.630 & 0.586 & \cellcolor[rgb]{ .806,  .802,  .802}0.640 & 0.440 & \cellcolor[rgb]{ .806,  .802,  .802}0.556 & 0.497 & 0.542 & 0.553 & 0.653 & 0.619 & \cellcolor[rgb]{ .806,  .802,  .802}0.664 \\
              & OV    & 0.632 & 0.670 & 0.674 & \cellcolor[rgb]{ .806,  .802,  .802}0.715 & 0.372 & 0.533 & 0.467 & \cellcolor[rgb]{ .806,  .802,  .802}0.533 & 0.556 & \cellcolor[rgb]{ .806,  .802,  .802}0.701 & 0.653 & 0.685 \\
              & BC    & 0.474 & 0.475 & 0.436 & \cellcolor[rgb]{ .806,  .802,  .802}0.489 & 0.407 & \cellcolor[rgb]{ .806,  .802,  .802}0.470 & 0.411 & 0.444 & 0.473 & \cellcolor[rgb]{ .806,  .802,  .802}0.587 & 0.512 & 0.526 \\
              & IV    & 0.546 & 0.594 & 0.586 & \cellcolor[rgb]{ .806,  .802,  .802}0.644 & 0.447 & \cellcolor[rgb]{ .806,  .802,  .802}0.541 & 0.521 & 0.486 & 0.550 & \cellcolor[rgb]{ .806,  .802,  .802}0.674 & 0.623 & 0.664 \\
              & VC    & 0.654 & 0.743 & 0.681 & \cellcolor[rgb]{ .806,  .802,  .802}0.746 & 0.443 & 0.575 & 0.512 & \cellcolor[rgb]{ .806,  .802,  .802}0.586 & 0.587 & 0.735 & 0.683 & \cellcolor[rgb]{ .806,  .802,  .802}0.744 \\
              & CM    & 0.668 & \cellcolor[rgb]{ .806,  .802,  .802}0.748 & 0.713 & 0.735 & 0.440 & \cellcolor[rgb]{ .806,  .802,  .802}0.587 & 0.514 & 0.573 & 0.606 & \cellcolor[rgb]{ .806,  .802,  .802}0.737 & 0.699 & 0.734 \\
              & SO    & 0.714 & \cellcolor[rgb]{ .806,  .802,  .802}0.703 & 0.625 & 0.647 & 0.554 & \cellcolor[rgb]{ .806,  .802,  .802}0.681 & 0.568 & 0.639 & 0.650 & 0.691 & 0.671 & \cellcolor[rgb]{ .806,  .802,  .802}0.724 \\
        \bottomrule[1.5pt]
        \end{tabular}%
    	\end{threeparttable}
     \label{tab:att_123}%
    \end{table*}%

    \begin{figure*}[!t]
    \centering
    \includegraphics[width=2\columnwidth]{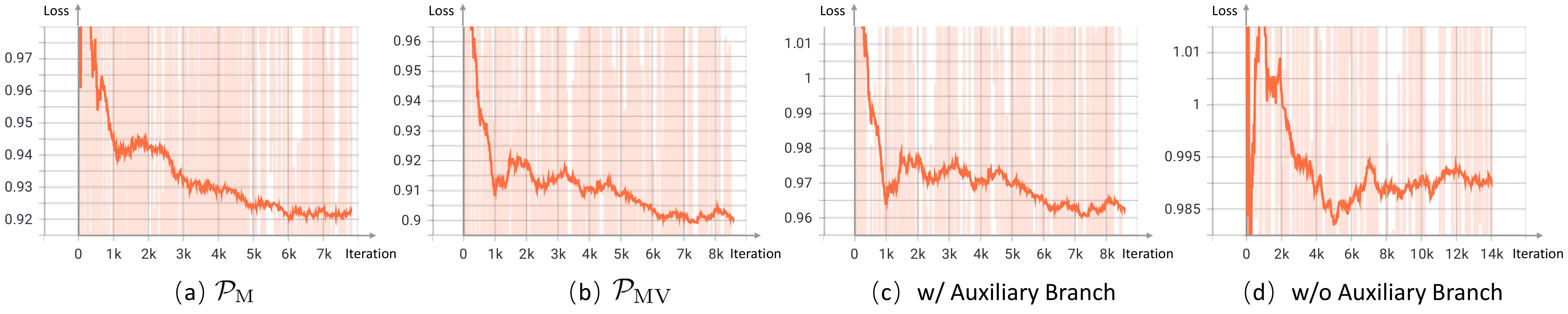}
    \caption{Training loss curves of PVT++ models. Coupled with visual feature, $\mathcal{P}_{\mathrm{MV}}$ can better learn to predict than $\mathcal{P}_{\mathrm{M}}$, thus the loss is observed to be smaller. Without auxiliary branch, the loss curve is less smooth, indicating the importance of $\mathcal{A}$.}
    \label{fig:loss}
\end{figure*}

\begin{table*}[!t]
\centering
\setlength{\tabcolsep}{1.2mm}
\fontsize{8}{9}\selectfont
\caption{Effect of extra latency brought by PVT++ in UAVDT~\cite{Du2018UAVDT} dataset. Here, the base tracker takes SiamRPN++$_{\mathrm{M}}$, whose original latency is fixed to $44.5$ ms/frame (its average onboard latency). We use $\cdot^\dagger$ to indicate neglecting the latency. With $\sim$5ms/frame extra time, the performance is slightly lower (2$\sim$3\% performance drop), while it is acceptable and still brings upto $15$\% performance gain.}
\begin{threeparttable}
\begin{tabular}{cccc|ccc|ccc}
\toprule[1.5pt]
Model & \multicolumn{3}{c|}{Tracker} & \multicolumn{3}{c|}{Tracker+$\mathcal{P}^\dagger_{\mathrm{MV}}$} & \multicolumn{3}{c}{Tracker+$\mathcal{P}_{\mathrm{MV}}$} \\
\midrule
Metric & mAUC$_{\Delta\%}$ & mDP$_{\Delta\%}$ & Latency & mAUC$_{\Delta\%}$ & mDP$_{\Delta\%}$ & Latency & mAUC$_{\Delta\%}$ & \multicolumn{1}{c}{mDP$_{\Delta\%}$} & Latency \\
Result & 0.494$_{+0.00}$ & 0.719$_{+0.00}$ & 44.5ms & 0.587$_{+18.8}$ & 0.825$_{+14.7}$ & 44.5ms & 0.576$_{+16.6}$ & 0.807$_{+12.2}$ & 50.0ms \\
\bottomrule[1.5pt]
\end{tabular}%
\end{threeparttable}
\label{tab:extra_latency}%
\end{table*}%

\begin{table*}[!t]
		\centering
		\setlength{\tabcolsep}{1.2mm}
		\fontsize{8}{9}\selectfont
		\caption{Performance of PVT++ models trained with different datasets. Full denotes $\sim$9,000 videos from VID~\cite{Russakovsky2015VID}, LaSOT~\cite{Fan2019LaSOTAH}, and GOT-10k~\cite{Huang2019GOT10kAL}. VID indicates using only $\sim$3,000 videos from VID~\cite{Russakovsky2015VID}. AVG means average results on the four test datasets. Since PVT++ utilizes the trained tracking models, We observe the training are not very sensitive to the scale of training set.}
		\begin{threeparttable}
			\begin{tabular}{cc|cc|cc|cc|cc|cc}
				\toprule[1.5pt]
				\multicolumn{2}{c|}{Dataset} & \multicolumn{2}{c|}{DTB70} & \multicolumn{2}{c|}{UAVDT} & \multicolumn{2}{c|}{UAV20L} & \multicolumn{2}{c|}{UAV123} & \multicolumn{2}{c}{AVG} \\
				\midrule
				PVT++ & Training & mAUC  & mDP   & mAUC  & mDP   & mAUC  & mDP   & mAUC  & mDP   & mAUC  & mDP \\
				\multirow{2}[0]{*}{$\mathcal{P}_\mathrm{V}$} & Full  & 0.352 & 0.472 & 0.564 & 0.799 & 0.488 & 0.675 & 0.504 & 0.703 & \cellcolor[rgb]{ .806,  .802,  .802}\textbf{0.477} & 0.662 \\
				& VID   & 0.362 & 0.483 & 0.519 & 0.752 & 0.497 & 0.694 & 0.513 & 0.731 & 0.473 & \cellcolor[rgb]{ .806,  .802,  .802}\textbf{0.665} \\
				\multirow{2}[1]{*}{$\mathcal{P}_\mathrm{MV}$} & Full  & 0.399 & 0.536 & 0.576 & 0.807 & 0.508 & 0.697 & 0.537 & 0.741 & \cellcolor[rgb]{ .806,  .802,  .802}\textbf{0.505} & \cellcolor[rgb]{ .806,  .802,  .802}\textbf{0.695} \\
				& VID   & 0.405 & 0.554 & 0.53  & 0.757 & 0.511 & 0.701 & 0.534 & 0.745 & 0.495 & 0.689 \\
				\bottomrule[1.5pt]
			\end{tabular}%
   \vspace{-0.4cm}
		\end{threeparttable}
  \label{tab:training_set}%
\end{table*}%
    
    We present full attribute-based analysis in \tref{tab:att_DTB}, \tref{tab:att_DT}, \tref{tab:att_20L}, and \tref{tab:att_123}. Following the previous work~\cite{Li2017DTB70}, we report results on aspect ratio variation (ARV), background clutter (BC), deformation (DEF), fast camera motion (FCM), in-plane rotation (IPR), motion blur (MB), occlusion (OCC), out-of-plane rotaTion (OPR), out-of-view (OV), scale variation (SV), and similar object around (SOA) in \tref{tab:att_DTB}. As shown in \tref{tab:att_DT}, results on background clutter (BC), camera rotation (CR), object rotation (OR), small object (SO), illumination variation (IV), object blur (OB), scale variation (SV), and large occlusion (LO), are reported for UAVDT~\cite{Du2018UAVDT}. For UAV20L and UAV123~\cite{Mueller2016UAV123}, we present results on scale variation (SV), aspect ratio change (ARC), low resolution (LR), fast motion (FM), full occlusion (FOC), partial occlusion (POC), out-of-view (OV), background clutter (BC), illumination variation (IV), viewpoint change (VC), camera motion (CM), and similar object (SO) in \tref{tab:att_20L} and \tref{tab:att_123}.
    
    We observe that the two modalities has their own advantage in different UAV tracking challenges. 
    For example, consider UAVDT dataset \cite{Du2018UAVDT} (\tref{tab:att_DT}), the visual branch is relatively good at challenges like camera rotation (CR), object rotation (OR), and scale variation (SV), where the object motion could be very complex and the visual appearance is helpful in prediction. 
    On the other hand, motion cues are robust when the visual feature is not reliable, for instance, similar object (SO) and large occlusion (LO) challenge. 
    In general, motion predictor is better than visual predictor, from which we conclude that past motion is still the main cue to inference future motion.
    While for the challenging UAV tracking, where motion could be extremely random and dynamic, introducing visual cues can significant improve the prediction robustness.
    Together, the jointly optimized model $\mathcal{P}_{\mathrm{MV}}$ is the most reliable for UAV latency-aware vsiaul tracking.

\section{Training Visualization}\label{app:training_visual}
    
    The training loss curves of PVT++ models with SiamRPN++$_{\mathrm{M}}$~\cite{Li2019SiamRPNEO} is shown in \fref{fig:loss}. Compared with motion predictor $\mathcal{P}_{\mathrm{M}}$, the joint predictor $\mathcal{P}_{\mathrm{MV}}$ can better learn to predict, resulting in smaller training loss. We also compared the losses from models with (c) or without (d) the auxiliary branch $\mathcal{A}$. Without $\mathcal{A}$, the loss curve fluctuates a lot, indicating that the model can't converge very well.

\section{Effect of Extra Latency}\label{app:extra_latency}
	
    PVT++ will bring a bit extra latency during online perception, which is negative for the performance. As shown in \tref{tab:extra_latency}, the latency of original tracker~\cite{Li2019SiamRPNEO} is about 45 ms/frame. Ignoring the predictor's latency, the online performance can reach 0.587 mAUC and 0.825 mDP. Taking the extra latency of $\sim$ 5 ms/frame into account, the result will slightly suffer, decreasing to 0.576 mAUC and 0.807 mDP. Therefore, though PVT++ introduces extra latency, the online performance can still be significantly improved by more than \textbf{10}\%. 

\section{Training Set Analysis}\label{app:training_set}
    
    Since PVT++ models can make full use of a trained tracker model, we find $\mathcal{P}_\mathrm{V}$ and $\mathcal{P}_\mathrm{MV}$ not very sensitive to the scale of training set. As shown in \tref{tab:training_set}, trained with only $\sim$3,000 videos from VID~\cite{Russakovsky2015VID}, our PVT++ can still converge well and achieve on par performance compared with the fully trained models.

\section{More Real-World Tests}\label{app:real_world}

	\begin{figure*}[!t]
		\centering
		\includegraphics[width=2\columnwidth]{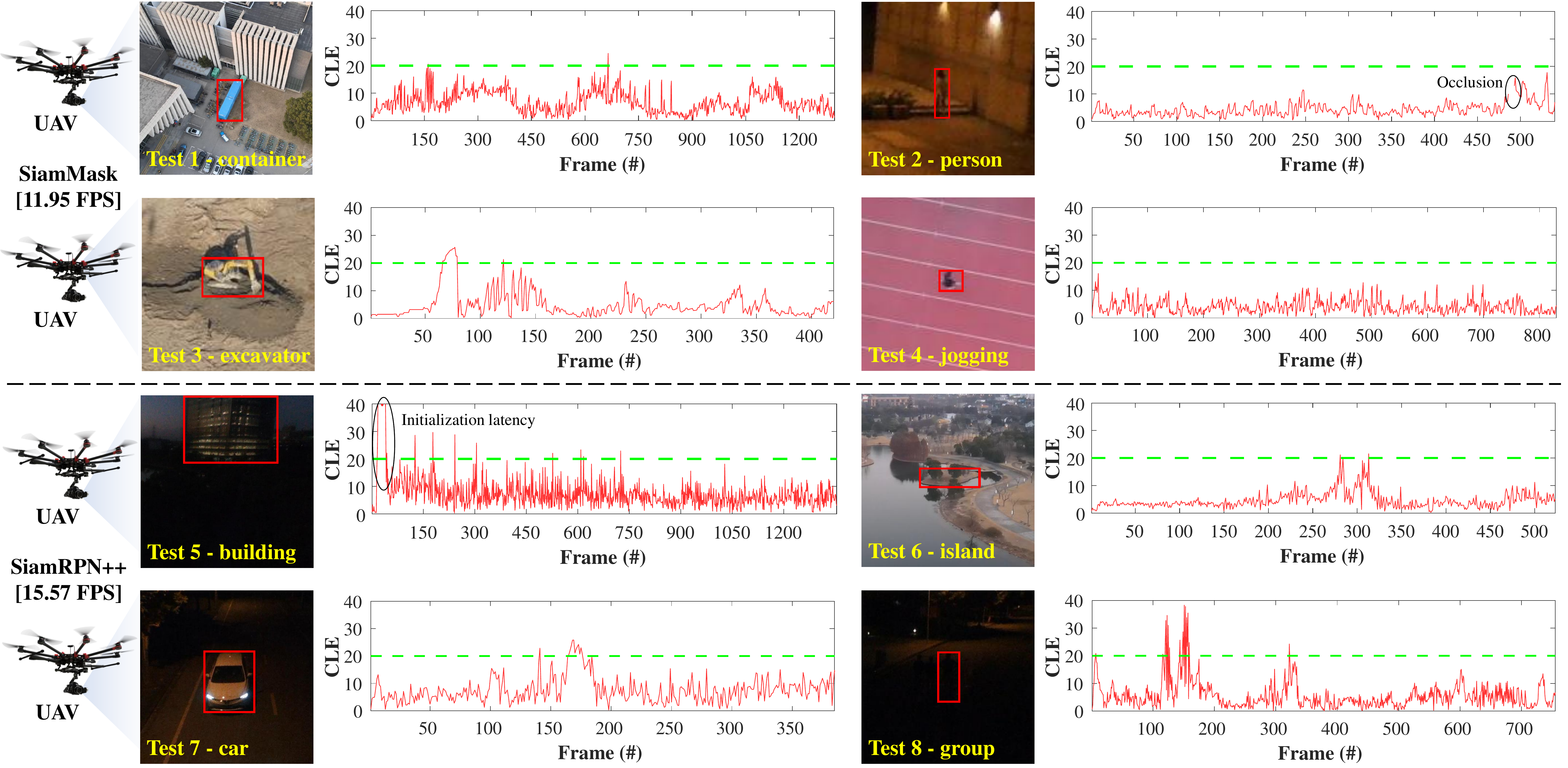}
		\caption{Eight real-world tests of PVT++ on non-real-time trackers, SiamMask \cite{Wang2019Mask} and SiamRPN++$_\mathrm{M}$ \cite{Li2019SiamRPNEO}. We present the tracking scenes, the target objects, and center location error (CLE) in the figure. Under various challenges like aspect ration change, illumination variation, low resolution, PVT++ maintains its robustness, with CLE below 20 pixels in most frames.}
		\vspace{-0.1cm}
		\label{fig:real_world}
	\end{figure*}

    In addition to the four real-world tests in Sec.~6.5 of the main paper, we present six more tests (together eight tests) in \fref{fig:real_world}, where we implemented the models on a real UAV and performed several flights. 
    The real-world tests involve two non-real-time trackers, SiamRPN++$_\mathrm{M}$ \cite{Li2019SiamRPNEO} ($\sim$ 15.57 FPS in the tests) and SiamMask \cite{Wang2019Mask} ($\sim$ 11.95 FPS in the tests), which are largely affected by their high onboard latency.
    Coupled with our PVT++ ($\mathcal{P}_{\mathrm{MV}}$), the predictive models work well under various tracking scenes, \textit{e.g.}, aspect ratio change in Test 1, dark environment in Test 2, 5, 7, and 8, view point change in Test 3, and occlusion in Test 2.
    The real-world tests also cover various target objects like person, building, car, and island, as shown in \fref{fig:real_world}. We have made them into videos for clear reference.
    The robustness of PVT++ in the onboard tests validate its effectiveness in the real-world UAV tracking challenges. 

    %%%%%%%%% REFERENCES
% {\small
% \bibliographystyle{ieee_fullname}
% \bibliography{egbib}
% }

\end{document}